\documentclass[10pt,twocolumn,letterpaper]{article}

\usepackage[pagenumbers]{cvpr} 

\usepackage{bm}
\usepackage{graphicx}
\usepackage{amsmath}
\usepackage{amssymb}
\usepackage{booktabs}
\usepackage[font={small}]{caption}

\usepackage{cuted}
\usepackage{capt-of}

\usepackage[misc]{ifsym}

\usepackage[pagebackref,breaklinks,colorlinks]{hyperref}
\newcommand{\qianyi}[1]{{\color{red}[qianyi: #1]}}
\newcommand{\chengyao}[1]{{\color{cyan}[chengyao: #1]}}
\newcommand{\wayne}[1]{{\color{blue}[wayne: #1]}}

\newcommand{\misscite}{\textcolor{red}{[C]~}}
\newcommand{\missref}{\textcolor{red}{[R]~}}

\newcommand\blfootnote[1]{%
  \begingroup
  \renewcommand\thefootnote{}\footnote{#1}%
  \addtocounter{footnote}{-1}%
  \endgroup
}

\usepackage[capitalize]{cleveref}
\crefname{section}{Sec.}{Secs.}
\Crefname{section}{Section}{Sections}
\Crefname{table}{Table}{Tables}
\crefname{table}{Tab.}{Tabs.}

\def\cvprPaperID{1573} 
\def\confName{CVPR}
\def\confYear{2022}

\begin{document}


\title{\vspace{-0.35cm}TransEditor: Transformer-Based Dual-Space GAN for\\Highly Controllable Facial Editing\vspace{-0.3cm}}

\author{
Yanbo Xu$^{1,4*}$ \hspace{12pt}
Yueqin Yin$^{1*}$ \hspace{12pt} 
Liming Jiang$^{2}$ \hspace{12pt}
Qianyi Wu$^{5}$ \\[2pt]
Chengyao Zheng$^{3}$ \hspace{12pt}
Chen Change Loy$^{2}$\hspace{12pt}
Bo Dai$^{2}$ \hspace{12pt}
Wayne Wu$^{1,3}$\textsuperscript{\Letter}\\[2pt]
$^1$Shanghai AI Laboratory \hspace{12pt}
$^2$S-Lab, Nanyang Technological University \hspace{12pt}
$^3$SenseTime Research\\[1pt]
$^4$Hong Kong University of Science and Technology\hspace{12pt}
$^5$Monash University\hspace{12pt}\\[1pt]
{\tt\small yxubu@connect.ust.hk} \hspace{12pt}
{\tt\small yinyueqin0314@gmail.com} \hspace{12pt}
{\tt\small qianyi.wu@monash.edu} \\[1pt]
{\tt\small \{liming002, ccloy, bo.dai\}@ntu.edu.sg} \hspace{12pt}
{\tt\small zhengchengyao@sensetime.com} \hspace{12pt}
{\tt\small wuwenyan0503@gmail.com}\vspace{-0.15cm}
}

\twocolumn[{
  \renewcommand\twocolumn[1][]{#1}
  \maketitle
  \begin{center}
  \vspace{-0.7cm}
  \includegraphics[width=0.95\textwidth]{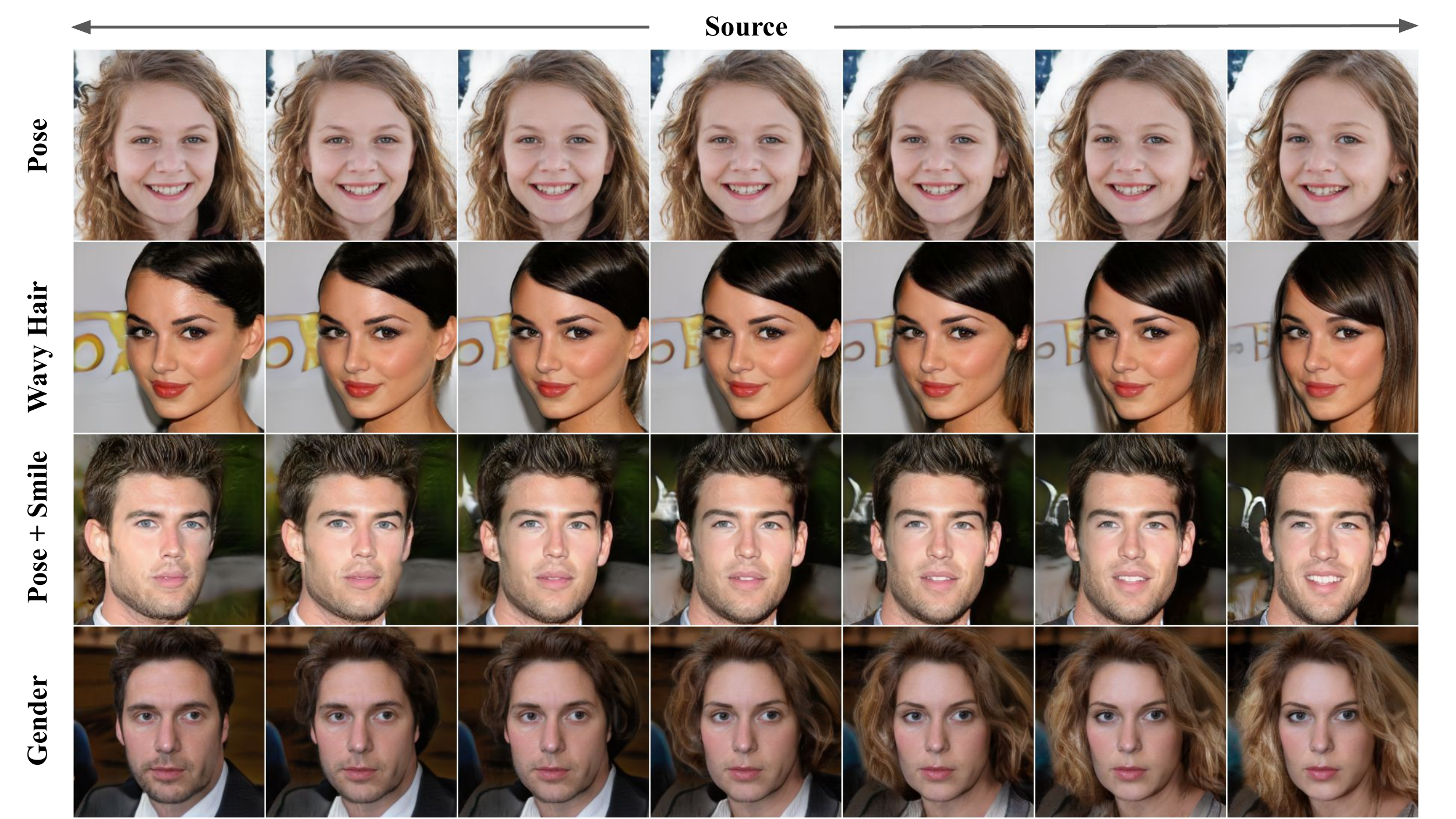}
  \vspace{-0.25cm}
  
   \captionof{figure}{\textbf{Diverse editing examples by TransEditor.} Images on the fourth column are the sampled source images, which are semantically interpolated to the left and right sides. With our proposed dual latent spaces, namely $\mathcal{P}$-space and $\mathcal{Z}$-space, different attributes can be edited in one space alone or in dual spaces simultaneously, demonstrating the flexibility of our method. Head pose (the first row) and wavy hair (the second row) are edited via the $\mathcal{P}$-space alone. Disentangled dual spaces enable the editing of pose and smile simultaneously (the third row) by editing pose in the $\mathcal{P}$-space and smile in the $\mathcal{Z}$-space. The Cross-Space Interaction allows the cooperated editing of complex attributes like gender (the last row).} 
  
  
  \vspace{-0.1cm}
  \label{fig:teaser}
  \end{center}
}]




\begin{abstract}
\vspace{-0.2cm}
Recent advances like StyleGAN have promoted the growth of controllable facial editing.
To address its core challenge of attribute decoupling in a single latent space, attempts have been made to adopt dual-space GAN for better disentanglement of style and content representations.
Nonetheless, these methods are still incompetent to obtain plausible editing results with high controllability, especially for complicated attributes.
In this study, we highlight the importance of interaction in a dual-space GAN for more controllable editing. We propose \textbf{TransEditor}, a novel Transformer-based framework to enhance such interaction.
Besides, we develop a new dual-space editing and inversion strategy to provide additional editing flexibility.
\blfootnote{\hspace{-0.2cm}$^*$ Two authors have equal contributions, ordered alphabetically. This work was done during an internship at Shanghai AI Laboratory.}
Extensive experiments demonstrate the superiority of the proposed framework in image quality and editing capability, suggesting the effectiveness of TransEditor for highly controllable facial editing. Code and models are publicly available at \url{https://github.com/BillyXYB/TransEditor}.

\end{abstract}


\section{Introduction}

High-fidelity generative modeling has made a dramatic leap thanks to the progress of Generative Adversarial Networks (GANs)~\cite{goodfellow2014generative,karras2019style,karras2020analyzing}.
These efforts have expedited the advancement in facial editing~\cite{zhang2018generative, he2019attgan, gu2020image}, an important downstream task with many practical applications, circumventing the cumbersome manual editing processes.
Nevertheless, \textit{highly controllable} facial attribute editing remains challenging when applying current approaches~\cite{karras2020analyzing,kim2021exploiting, kwon2021diagonal}.

The main challenge of highly controllable facial editing lies in the clean disentanglement of attributes in the latent space. For instance, it is expected to maintain the consistent face characteristic when manipulating the head pose of a portrait.
Representative GAN-based methods~\cite{karras2020analyzing,kim2021exploiting} explored the single latent space representation and the style modulation technique of generators for better image generation and semantic editing.
On another note, some studies~\cite{wu2021stylespace,collins2020editing} focus on the local-region editing for specific facial parts in a fine-grained manner.
However, under the single latent space design, these works still suffer from the latent entanglement of certain complicated attributes, such as facial pose.

Working towards these challenges, recent studies~\cite{alharbi2020disentangled,kwon2021diagonal} have presented preliminary attempts that leverage the idea of dual latent spaces for style and content disentanglement within the StyleGAN~\cite{karras2019style} architecture, achieving plausible semantic separation in each space.
However, these attempts may not be adequate to take full advantage of the potential of a dual-space GAN. We empirically observed that these methods fail to obtain decent facial editing results, especially in complicated attributes, as shown in~\cref{fig:sample image editing}.
Altering the pose of a face may produce dissatisfactory artifacts, and a drastic change in hue easily occurs when interpolating the style code of DAT~\cite{kwon2021diagonal} by fixing the content code.
Accordingly, we conjecture that the underlying cause is the lack of \textit{interaction} between the two latent spaces, and thus editing one space may distort the other, making the controllable facial editing infeasible.

%
%

In this paper, we show that the latent space \textit{interaction} in a dual-space GAN plays a significant role in facial editing. Inspired by recent immense success of Transformers~\cite{dosovitskiy2021an,carion2020end,jiang2021transgan,wang2021end,li2021transforming} in vision tasks, we propose a novel Transformer-based dual-space GAN, named \textbf{TransEditor}, to strengthen the dual latent space interaction. 
The two latent spaces in our generator, \ie, $\mathcal{P}$-space and $\mathcal{Z}$-space, serve as the initial input feature map of the generator and the style modulation~\cite{karras2020analyzing} for all layers, respectively. 
Specifically, we propose a \textit{Transformer-Based Cross-Space Interaction} module, where we incorporate a cross-space attention mechanism based on the Transformer to enhance the interaction between these two spaces, facilitating highly controllable facial editing. 
The design of interaction is \textit{non-trivial} since potential entanglement might be aggravated via the interaction.
We propose to use the $\mathcal{P}$-space as the query and the $\mathcal{Z}$-space as the key and value.
Therefore, $\mathcal{P}$-space is only exploited to re-weight the value matrix from $\mathcal{Z}$-space through the Cross-Space Interaction module, hence still being disentangled with the refined code from $\mathcal{Z}$-space.
Thanks to the design of the interaction module, our dual spaces allow flexible editing while maintaining disentanglement semantically in an expressive manner, with $\mathcal{P}$-space mainly controlling the structural information and $\mathcal{Z}$-space determining the texture representation. Then, different from previous methods~\cite{Shen_2020_CVPR,shen2020interfacegan,shen2021closed} that perform 
editing in a single space, we devise a novel \textit{Dual-Space Editing} strategy to leverage the controllability brought by TransEditor (\cref{fig:teaser}). Further, to enable the real image editing, we also extend existing inversion techniques~\cite{richardson2021encoding} to fit the proposed dual-space design.

The contributions of this work can be summarized as follows. We propose TransEditor, a novel Transformer-based dual-space GAN for highly controllable facial attribute editing. Through the introduced cross-space attention mechanism, the two latent spaces can establish meaningful interaction in a disentangled manner. Additionally, we develop a novel flexible dual-space image editing and inversion strategy to leverage the improved controllability provided by TransEditor. Extensive experiments demonstrate the effectiveness of TransEditor in highly controllable and stable facial attribute editing, outperforming state-of-the-art approaches especially for complicated attributes.

\section{Related Work} \label{Related Work}

\subsection{Structured Latent Space for Disentanglement}
\noindent\textbf{Single Latent Space.} 
ProgressiveGAN~\cite{karras2017progressive} uses a latent code as the input feature map to control the generation of the whole image. StyleGAN~\cite{karras2019style, karras2020analyzing} improves the disentanglement  using a mapping network that maps the initial distribution to an intermediate latent space $w$, thus the distribution of latent codes can be reformed to the given real image distribution. Besides, the layer-wise Adaptive Normal Instance (AdaIN)~\cite{ghiasi2017exploring, huang2017arbitrary} is used to improve the disentanglement further. StyleMapGAN~\cite{kim2021exploiting} reshapes the latent code into a tensor with spatial structure. The method achieves good reconstruction results but poor semantic editing performance, known as the Editability-Distortion trade-off~\cite{alaluf2021restyle, tov2021designing, zhu2020improved}. 
Its encoder also needs to be trained together, the training will fail otherwise. The reason is that the network relies on the encoder to make the projected style map of the real images achieve local correspondence~\cite{kim2021exploiting}.


\noindent\textbf{Dual Latent Spaces.}
To achieve better disentanglement or spatial awareness, some studies that explore dual latent spaces design have been proposed. SNI~\cite{alharbi2020disentangled} separates the latent space into spatially-variable and spatially-invariable parts. The separation allows a certain degree of local editing. However, the limited capacity of the spatial code (4$\times$4 or 8$\times$8) may cause failure when performing content control~\cite{kwon2021diagonal}. DAT~\cite{kwon2021diagonal} improves the structure by introducing a symmetric space similar to the origin latent space and utilizing separate operations during the generating process for disentanglement. However, the content space proposed by DAT~\cite{kwon2021diagonal} operates on each pixel individually, so it lacks a global structure connection and cannot perform well on some attributes with large structural changes like pose. 
There are other flavors of dual latent spaces for various tasks. Zhu~\etal~\cite{zhu2021barbershop} propose $\mathcal{F}$-space and $\mathcal{S}$-space for image compositing. Park~\etal~\cite{park2020swapping} use a style code vector and a structure code with spatial dimension which is similar to SNI~\cite{alharbi2020disentangled}. ~\cite{lewis2021tryongan,albahar2021pose} disentangle the pose and style for virtual try-on.

\subsection{GAN-Based Image Editing}
\noindent\textbf{Latent Space Manipulation.}
Since the latent space of GAN is semantic-aware, it is possible to manipulate the attributes of the generated image through latent space navigation~\cite{zhu2020domain}. Previous works aiming at finding semantic directions can typically be divided into the supervised ones~\cite{Shen_2020_CVPR, goetschalckx2019ganalyze}, the self-supervised ones~\cite{gansteerability, spingarn2020gan}, and the unsupervised ones~\cite{harkonen2020ganspace, shen2021closed}. InterfaceGAN~\cite{Shen_2020_CVPR} finds hyper-planes in the latent space that corresponds to particular attributes by using attribute classifiers and SVM. Jahanian~\etal~\cite{gansteerability} need a large set of paired examples to fit the image transformations. SeFa~\cite{shen2021closed} performs decomposition on eigenvectors and finds interpretable directions without extra labels. Meanwhile, there are some studies~\cite{wu2021stylespace, collins2020editing} that focus on local editing for specific facial parts. StyleSpace~\cite{wu2021stylespace} proposes a space of channel-wise style parameters to perform highly localized and disentangled attribute editing.

\noindent\textbf{GAN Inversion.}
To edit a real image using Latent Space Manipulation, the image should be projected back into the latent space of the generator~\cite{zhu2020domain}. Optimization-based methods~\cite{ma2019invertibility, abdal2019image2stylegan} directly optimize the latent code to minimize the pixel-wise reconstruction loss. However, the optimization process is slow and the inverted code may land out of the original semantic manifold, making the editing process suffer from meaningful semantic manipulation. This problem is resolved by learning-based methods~\cite{richardson2021encoding, tov2021designing}, which use an extra encoder trained and guided by the generator to directly embed the image into a latent code, offering benefit during real image editing.


\subsection{Transformer-Based Interaction}
%
Multi-head attention module is often used in some multi-modality tasks to establish the inter-modal interaction between text modalities and visual modalities, such as image captioning~\cite{cornia2020meshed, huang2019attention} and text-to-image translation~\cite{ramesh2021zero, ding2021cogview}. TransStyleGAN~\cite{li2021transforming} introduces a Transformer structure to model the correlation between different layers of style codes within a single latent space based on the StyleGAN2 architecture. In this work, we leverage the multi-head attention module in the Transformer~\cite{vaswani2017attention} to establish the interaction between the proposed dual latent spaces (\ie, $\mathcal{Z}$-space and $\mathcal{P}$-space), facilitating more flexible and controllable attribute editing.

\section{Methodology}

\begin{figure*}[h!]
  \centering
    \vspace{-0.15cm}
    \includegraphics[width=\textwidth]{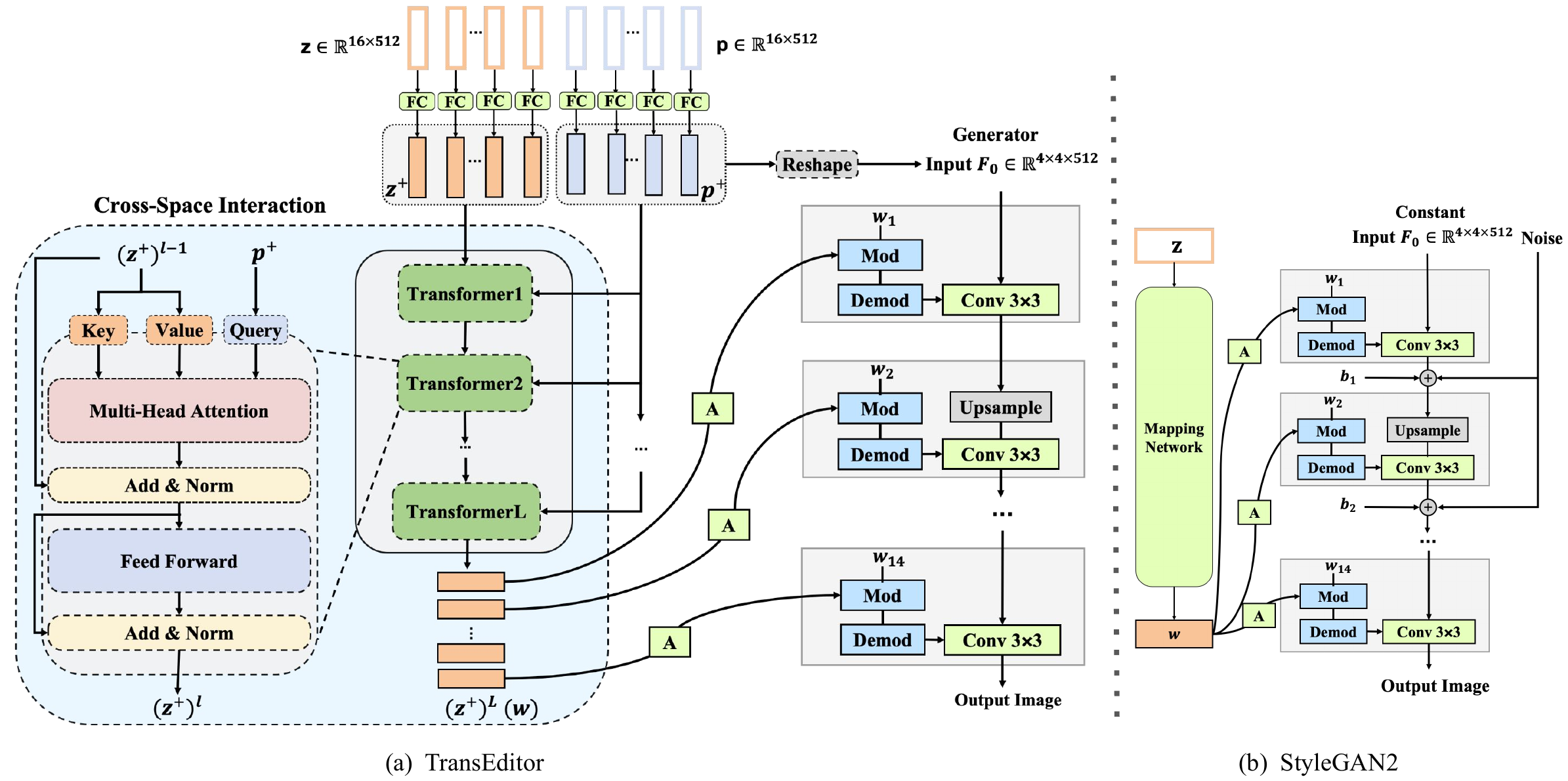}
     \vspace{-0.55cm}
    \caption{\textbf{Network architecture of TransEditor.} (a) shows the structure of our model, which contains two separate latent spaces $\mathcal{Z}$ and $\mathcal{P}$, a Cross-Space Interaction module based on the Transformer, and a generator. Compared to (b) StyleGAN2~\cite{karras2020analyzing} that leans a constant input, our generator uses the $\bm{p}^{+}$ code as the input and the interaction result $(\bm{z}^{+})^{L} (\bm{w})$ for style modulation.}
    \vspace{-0.25cm}
   \label{fig:model_architecture}
\end{figure*}

 Our goal is to achieve more controllable facial attribute editing through the disentangled but collaborative dual spaces. \cref{fig:model_architecture} (a) shows the structure of our model. We propose two latent spaces $\mathcal{P}$ and $\mathcal{Z}$ with separate mappings (\cref{dual-latent-spaces}), which are used as the initial input feature map of the generator and the layer-wise style modulation, respectively. Then, an interaction module based on the Transformer (\cref{Transformer-based Cross-space Interaction}) is proposed to model the interaction between these two spaces, making them more balanced during editing.
 Further, we develop a new dual-space image editing and inversion strategy (\cref{dual-space-editing}) for real image editing. 



\subsection{Dual Latent Spaces}\label{dual-latent-spaces}
 Instead of learning a generator $\mathbf{G}$ that maps a single Gaussian distribution to an image $\bm{x}$, \ie, $\bm{x}=\mathbf{G}(\bm{z}), \bm{z} \in \mathcal{N}(0, \mathbf{I})$, two separated latent spaces are used in our method, denoted as $\mathcal{Z}$ and $\mathcal{P}$. Thus, our generation process can be reformulated as: 
\begin{equation}
\bm{x} = \mathbf{TransEditor}(\bm{z},\bm{p}),
\end{equation} where $(\bm{z}, \bm{p}) \in \mathcal{Z} \times \mathcal{P}$. 

Note that it is \textit{non-trivial} to determine how the two spaces should be integrated with the generator.
For the StyleGAN2-based architecture\cite{karras2020analyzing}, at each layer, the generation process can be expressed as:
\begin{equation}
  \bm{F}_{i+1} = \mathbf{ModuConv}(\bm{T}_{\bm{w}_{i}}, \bm{F}_{i}),
  \label{eq:important}
\end{equation}
 where $\bm{F}_{i}$ denotes the feature map produced from the previous \(i-1\) layer, and $\bm{T}_{\bm{w}_{i}}$ is the modulation and demodulation process determined by the layer style code $\bm{w}_{i}$. 
Specifically, $\bm{T}_{\bm{w}_{i}}$ scales the parameters of the convolution module in layer $i$ by:
\begin{equation}
  \bm{T}_{\bm{w}_{i}}: w''_{ijk} = w'_{ijk} \bigg/ \sqrt{\raisebox{0mm}[4.0mm][2.5mm]{$\underset{j,k}{{}\displaystyle\sum{}}$} {w'_{ijk}}^2 + \epsilon},
\end{equation}
where \(w_{ijk}^\prime = s_{i} * w_{ijk}\), and $j,k$ are the convolution entries. $\bm{F}_{i}$ will then be convoluted by the scaled parameters $\bm{w}_{i}^{\prime \prime}$. Although the modulation and demodulation are performed at every layer, each feature map is the convoluted result from the previous layer. Thus, the initial feature map $\bm{F}_{0}$ is the foundation that the whole generation process is based on. 
Therefore, compared with StyleGAN2 in \cref{fig:model_architecture}(b), to provide more controllability, we replace the learned constant input 
with the latent input from $\mathcal{P}$-space. Besides, the additional latent capacity provided by $\mathcal{P}$-space allows us to remove the noise inputs.



In addition, we consider that reshaping a single sampled vector for the entire latent input is inherently entangled~\cite{alharbi2020disentangled}, hence our two spaces $\mathcal{Z}$ and $\mathcal{P}$ consist of separate sub-vectors, \ie, $\bm{z} \in \mathbb{R}^{n\times512}, \bm{p} \in \mathbb{R}^{n\times512}$. To further encourage a desirable disentanglement property, we exploit separate mappings for the dual spaces: $(\bm{z}^{+}, \bm{p}^{+}) \in M_{\bm{z}}(\mathcal{Z}) \times M_{p}(\mathcal{P}) $, which can be written as:
\begin{equation}
     \bm{z}^{+} = \begin{bmatrix}
                 \bm{z}^{+}_{1} \\
                 \bm{z}^{+}_{2} \\
                 ... \\
                 \bm{z}^{+}_{n}
             \end{bmatrix} = 
             \begin{bmatrix}
                 M_{\bm{z}_{1}} & 0 & ... & 0\\
                 0 & M_{\bm{z}_{2}} & ... & 0\\
                 ... \\
                 0 & 0 & ... & M_{\bm{z}_{n}}
             \end{bmatrix}
             \begin{bmatrix}
                 \bm{z}_{1} \\
                 \bm{z}_{2} \\
                 ... \\
                 \bm{z}_{n}
             \end{bmatrix}
 \end{equation}
 \begin{equation}
     \bm{p}^{+} = \begin{bmatrix}
                 \bm{p}^{+}_{1} \\
                 \bm{p}^{+}_{2} \\
                 ... \\
                 \bm{p}^{+}_{n}
             \end{bmatrix} = 
             \begin{bmatrix}
                 M_{\bm{p}_{1}} & 0 & ... & 0\\
                 0 & M_{\bm{p}_{2}} & ... & 0\\
                 ... \\
                 0 & 0 & ... & M_{\bm{p}_{n}}
             \end{bmatrix}
             \begin{bmatrix}
                 \bm{p}_{1} \\
                 \bm{p}_{2} \\
                 ... \\
                 \bm{p}_{n}
             \end{bmatrix}
 \end{equation}
Note that each \(M_{\bm{z}_{i}}\) or \(M_{\bm{p}_{i}}\) is a MLP~\cite{gardner1998artificial} module, and the mapped space $\bm{z}^{+}\times \bm{p}^{+}\in \mathbb{R}^{n\times512}\times \mathbb{R}^{n\times512}$. In our experiments, we set \(n=16\).

\subsection{Transformer-Based Cross-Space Interaction} \label{Transformer-based Cross-space Interaction}
In a dual-space GAN, naive generation using two separated latent codes might be problematic. 
SNI~\cite{alharbi2020disentangled} shows that adding the style code at all layers influences the disentanglement performance, \ie, changing the style code at early layers of the generator affects the structural information. Although DAT~\cite{kwon2021diagonal} achieves better disentanglement than SNI, we find that when fixing DAT's content code and interpolating its style code, drastic changes in hue could easily appear and cause artifacts.
We attribute this phenomenon to the lack of interaction in the dual latent spaces, as they are not correlated by any means.

Inspired by the cross-domain Transformer model~\cite{ramesh2021zero}, we correlate the two separated spaces via a cross-attention-based interaction module. The mapped latent code $\bm{z}^{+}$ is used as the key ($K$) and value ($V$), and the latent code $\bm{p}^{+}$ as the query ($Q$). The interaction in the $l$-th layer transformer can be written as:
\begin{gather}
    \emph{Q} = \bm{p}^{+}W^{Q},\emph{K} = (\bm{z}^{+})^l{W}^{K}, \emph{V} = (\bm{z}^{+})^l{W}^{V}, \\
    (\bm{z}^{+})^{l+1} = \mathbf{softmax}(\frac{\emph{Q}\emph{K}^\mathsf{T}}{\sqrt{d_{k}}})\emph{V} + (\bm{z}^{+})^{l}, \label{interation formula}
\end{gather}
where $W^{Q}, W^{K}, W^{V}$ are linear projection matrices, and \(d_k\) is the common dimensionality of the latent code. In other words, an attention process queried by $\bm{p}^{+}$ will be operated on $\bm{z}^{+}$. Since $\bm{p}^{+}$ acts as the query only, the two spaces are still separated, yet $\bm{z}^{+}$ has adopted to the query $\bm{p}^{+}$. This design enables better disentanglement of the dual spaces while maintaining global consistency during editing~(see~\cref{ablation_kqv}). 

The refined latent code $(\bm{z}^{+})^{L}$, that is $\bm{w}$, serves as the style modulation parameters of the generator $\mathbf{G}$.
The generation process of image $\bm{x}$ is thus formulated as:
\begin{gather}
    \bm{x} = \mathbf{G}(\bm{w}, \bm{F}_{0}),
\end{gather}
where $\bm{w}$ is the modulation input produced from~\cref{interation formula}, and the Reshape module (see~\cref{fig:model_architecture}(a)) reshapes the mapped $\bm{p}^{+}$ code into the initial feature map $\bm{F}_{0}$ whose spatial dimension is suitable as the input to the generator.
The training of our generator is entirely unsupervised, which only applies the adversarial loss and path length regularization following StyleGAN2~\cite{karras2020analyzing}.

\subsection{Dual-Space Image Editing and Inversion}\label{dual-space-editing}
Intuitively, some complicated attributes (\eg, age) may involve changes in both the facial structure and the texture. With our disentangled dual latent spaces, we propose to edit such complex attributes using $\mathcal{Z}$ and $\mathcal{P}$ together. To our knowledge, this is the first study that performs attribute editing through the two latent spaces simultaneously.

As enforcing the semantic distribution to fit the original Gaussian distribution is undesirable~\cite{karras2019style}, the editing procedure of our model via the latent manipulation is operated on the product space $\mathcal{Z}^{+} \times \mathcal{P}^{+}$. Hence, our dual-space manipulation procedure can be written as:
\begin{equation}
    I_{(\mathcal{Z}^{+} \times P^{+})}(\bm{z}^{+}, \bm{p}^{+}) = (I_{\mathcal{Z}^{+}}(\bm{z}^{+}), I_{\mathcal{P}^{+}}(\bm{p}^{+})),
\end{equation}
where $\bm{z}^{+} \in \mathcal{Z}^{+}, \bm{p}^{+} \in \mathcal{P}^{+}$, and \(I_{i}\) represents the manipulation operated on space \(i\).

It is noteworthy that the two operations $I_{\mathcal{Z}^{+}}$ and $I_{\mathcal{P}^{+}}$ could be different. In our case, we utilize the linear latent manipulation method by InterFaceGAN~\cite{shen2020interfacegan} for both spaces. Specifically, for each attribute, we train two hyper-planes in the two separated latent spaces using an SVM classifier, thus obtain the normal vector $n_z$ in $\mathcal{Z}^{+}$-space and normal vector $n_p$ in $\mathcal{P}^{+}$-space. Thereafter, for sampled latent codes $\bm{z}^{+}$ and $\bm{p}^{+}$, we can move $\lambda_{z}$ steps along $n_z$ and $\lambda_{p}$ steps along $n_p$ to get the new latent codes $(\bm{z}^{+} + \lambda_{z} * n_z, \bm{p}^{+} + \lambda_{p} * n_p)$. For the attributes that are fully contained in one space only, we edit them by operating on that space solely(\eg, $\lambda_p = 0$). More complicated attributes like gender will be better manipulated if the two spaces are used simultaneously. The manipulated codes produced from $I_{(\mathcal{Z}^{+} \times \mathcal{P}^{+})}(\bm{z}^{+},\bm{p}^{+})$ are then be used for generating the edited image. 

\begin{figure}[]
  \centering
    \vspace{-0.18cm}
    \includegraphics[width=0.95\linewidth]{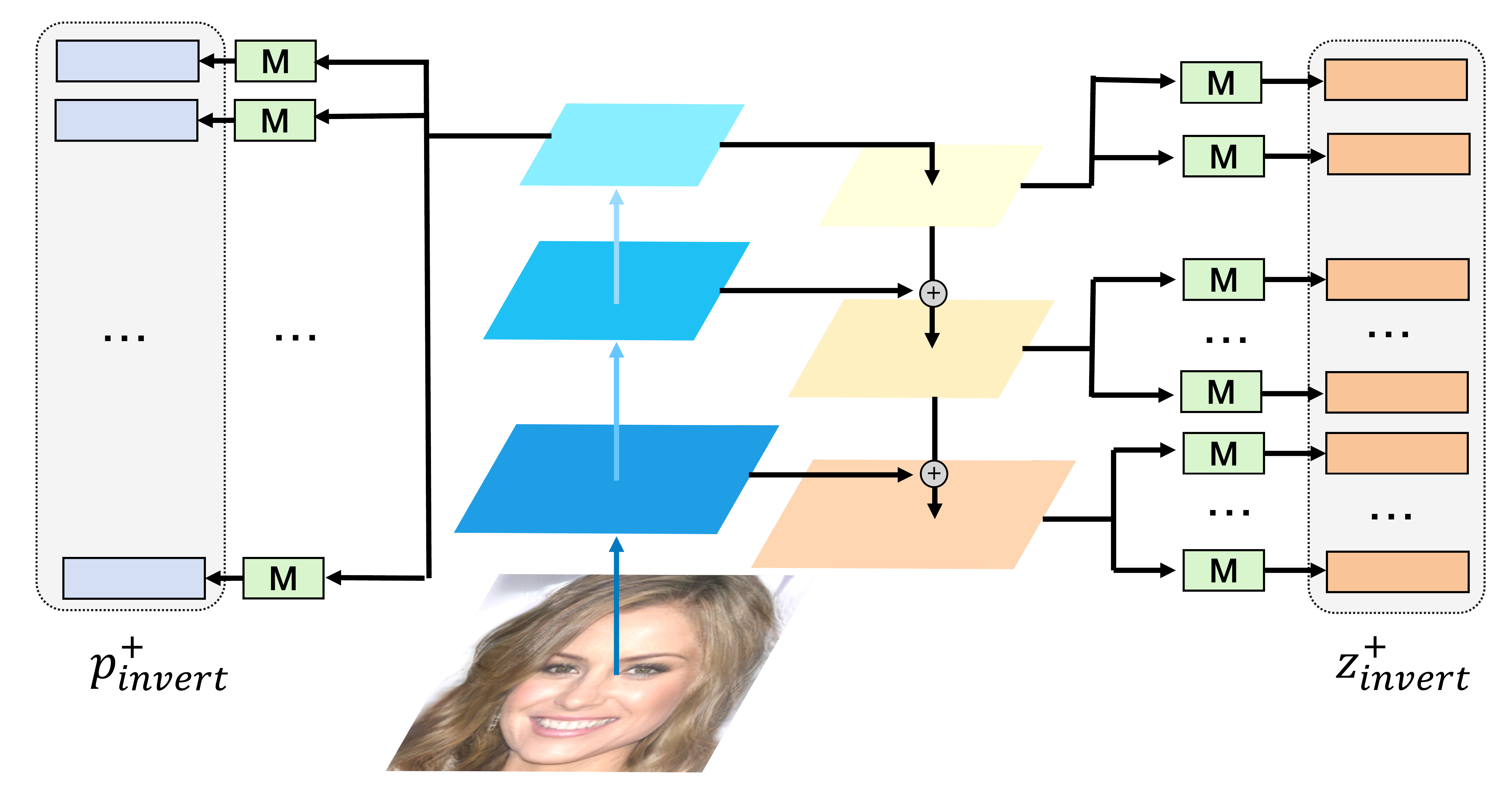}
    \vspace{-0.25cm}
    \caption{\textbf{Dual-Space Inversion Architecture.} The inverted $\bm{z}^{+}$ code comes from different layers of the extracted feature map, while $\bm{p}^{+}$ code comes from the highest feature map.}
    \vspace{-0.25cm}
   \label{fig:inversion_method}
\end{figure}
To edit real images, it is necessary to invert the images back into the dual latent spaces. We adopt a pSp~\cite{richardson2021encoding} encoder architecture for our dual-space image inversion.
As shown in~\cref{fig:inversion_method}, three levels of feature maps of the input real image are first extracted using a feature pyramid.
Since our $\bm{z}^{+}$ space has a layer-wise structure, different features are used to produce each $\bm{z}^{+}_{i}$.
The $\bm{p}^{+}$ latent code is only mapped from the highest-level feature in the encoder and injected as the initial feature map input of the generator.

The aforementioned inversion strategy maps real images into our trained dual latent spaces, and thereby we can apply linear latent manipulation to perform dual-space editing.

\begin{figure*}[!h]
    \begin{subfigure}[b]{0.49\textwidth}
        \includegraphics[width=\linewidth]{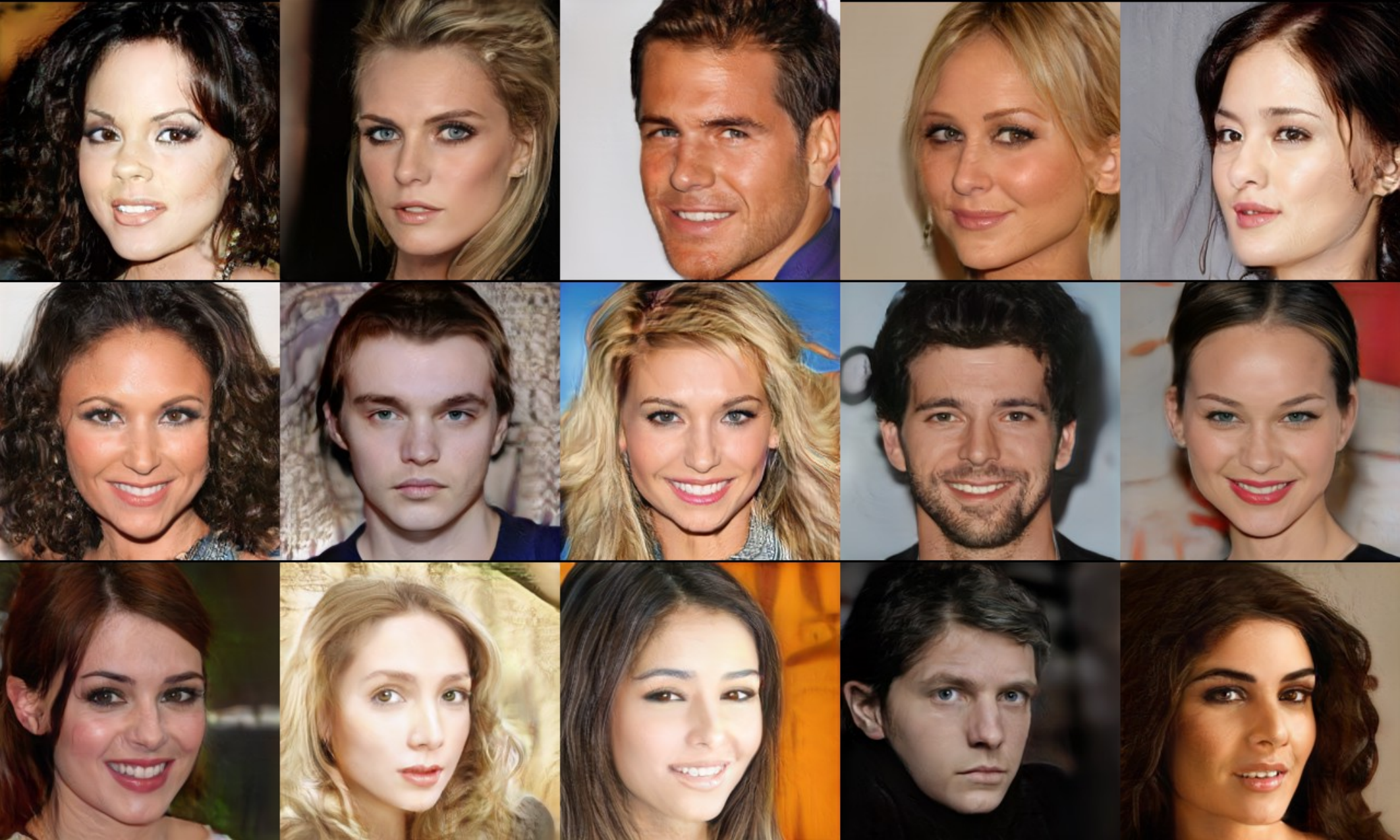}
            \caption{Fix the $\bm{p}$ code and sample $\bm{z}$ code}
            \label{fig:swap-celeb-z}
        \end{subfigure}%
        \hfill
        \begin{subfigure}[b]{0.49\textwidth}
        \includegraphics[width=\linewidth]{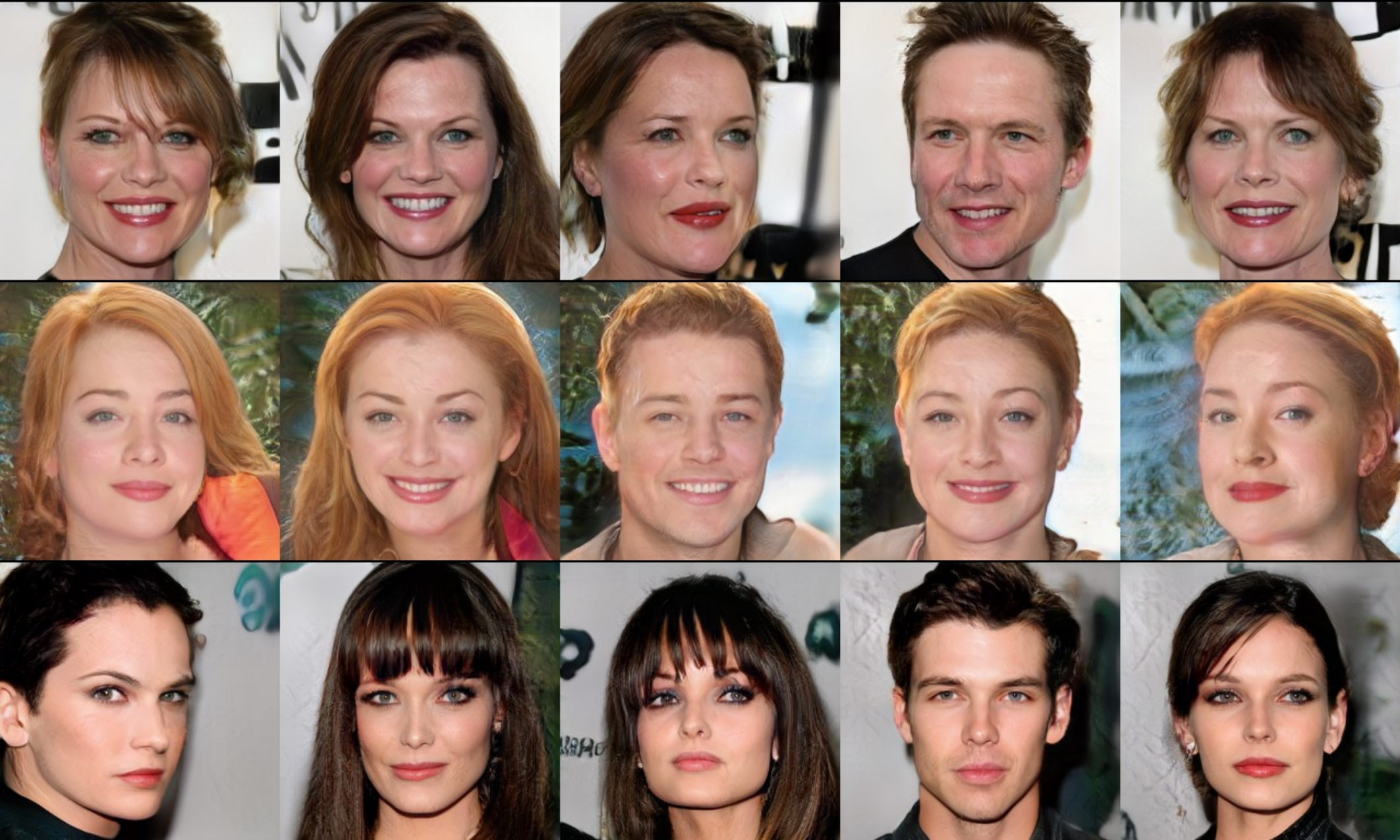}
            \caption{Fix the $\bm{z}$ code and sample $\bm{p}$ code}
    \label{fig:swap-celeb-p}
        \end{subfigure}%
        \hfill
        \begin{subfigure}[b]{0.49\textwidth}
                \includegraphics[width=\linewidth]{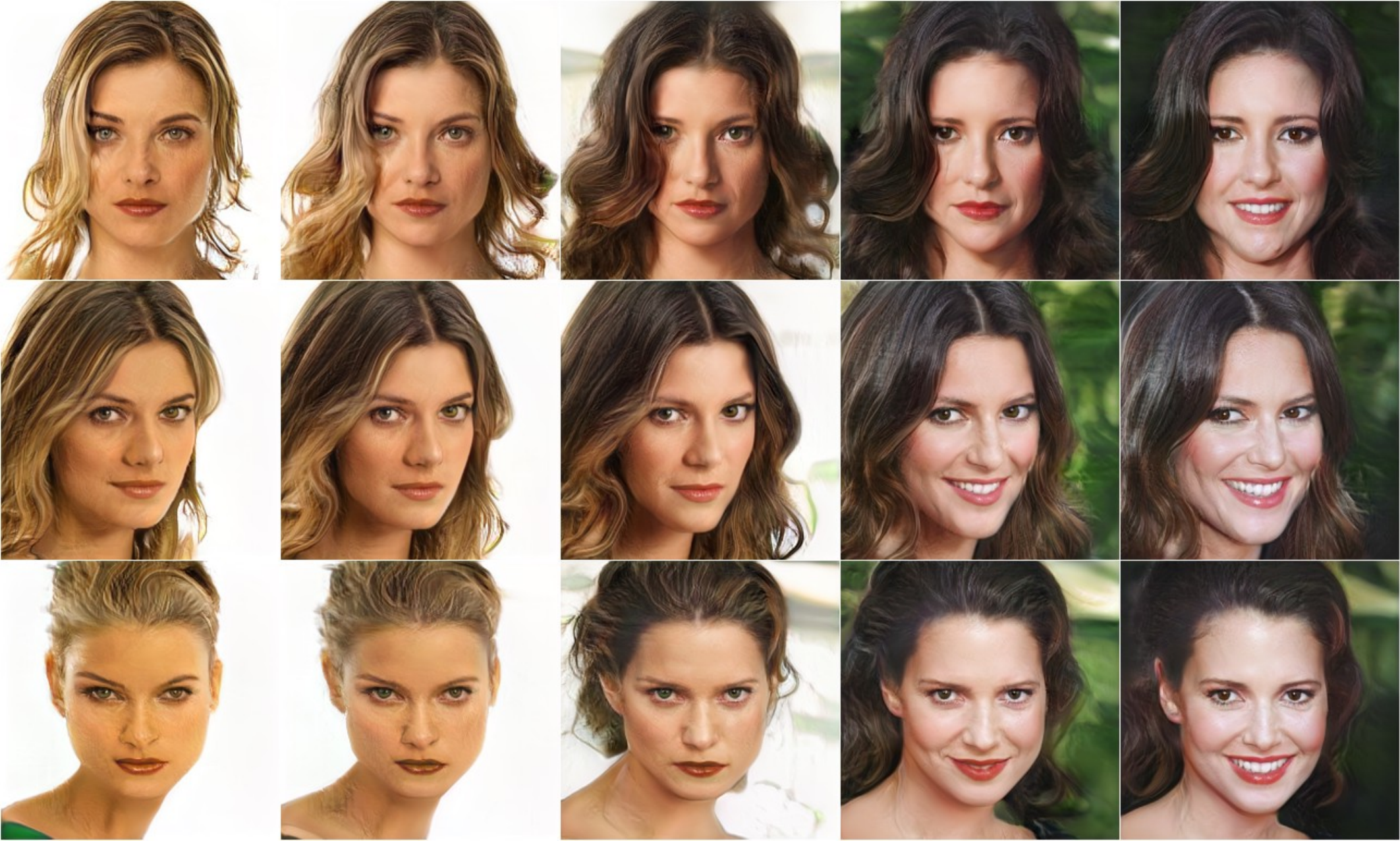}
                \caption{Interpolate $\bm{z}$ code}
    \label{fig:interp-celeb-z}
        \end{subfigure}%
        \hfill
        \begin{subfigure}[b]{0.49\textwidth}
                \includegraphics[width=\linewidth]{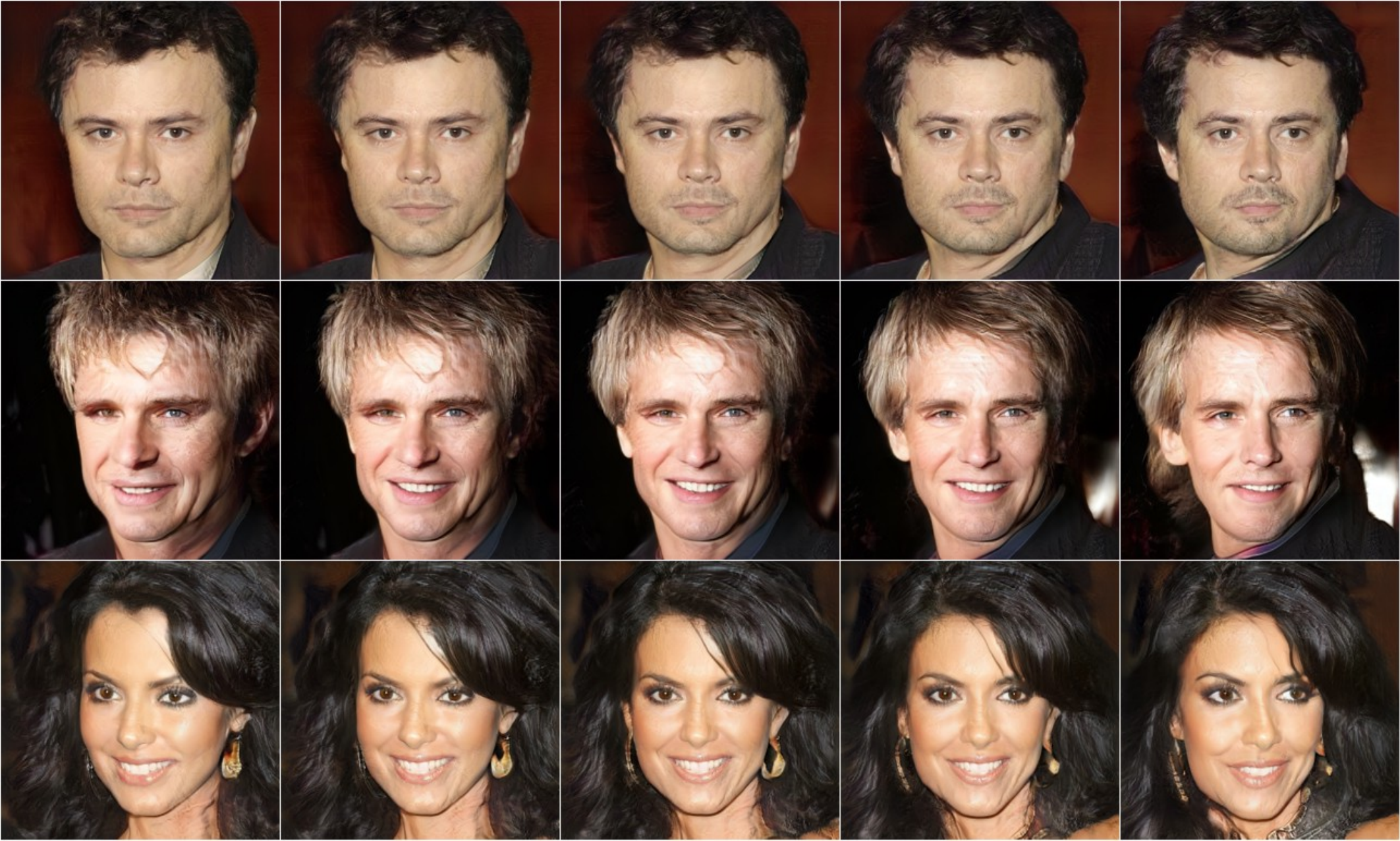}
                \caption{Interpolate $\bm{p}$ code}
    \label{fig:interp-celeb-p}
        \end{subfigure}
        \vspace{-0.25cm}
        \caption{\textbf{Dual Latent Spaces of TransEditor}. Each row in (a) is generated by a fixed $\bm{p}$ code and a randomly sampled $\bm{z}$ code. 
        Similarly, each row in (b) is generated by a fixed $\bm{z}$ code and a randomly sampled $\bm{p}$ code. In (c), each column starts from the same $\bm{z}$ code and interpolates towards the same direction. Each row has the same sampled $\bm{p}$ code. 
        Similarly, each column in (d) starts from the same $\bm{p}$ code then interpolates in the same way. Each row in (d) shares the same z code. 
        }\label{fig:swap-interp-celeba}
        \vspace{-0.25cm}
\end{figure*}
\section{Experiments}

In our experiments, we first evaluate the effectiveness of our method in highly controllable facial editing (Sec.~\ref{self_evaluation}). 
Then, we compare our mothod with three representative state-of-the-art methods (Sec.~\ref{comparion_to_sota}), \ie, the single-space methods (StyleGAN2~\cite{karras2020analyzing}, StyleMapGAN~\cite{kim2021exploiting}), and the dual-space method (DAT~\cite{kwon2021diagonal}), on both qualitative (\cref{qualitative}) and quantitative (\cref{quantitative}) aspects. In addition, we perform an ablation study (Sec.~\ref{ablation_study}) to isolate each pivotal components of our method. We trained TransEditor on CelebA-HQ~\cite{karras2017progressive} and FFHQ~\cite{karras2019style} at a resolution of $256 \times 256$. Details of experiment settings and implementation can be found in our appendix. The quantitative metrics used in this section are shown as follows.


\noindent\textbf{Re-scoring Calculation.}
We designed this metric given by \(C_{e}/C_{i}\) to quantitatively evaluate the editing performance, where $C_{e}$ and $C_{i}$ denote the cumulative change of attribute score of the edited and the influenced attributes, respectively. It effectively measures how the editing of one attribute affects other attributes. Details of this metric are included in the appendix. A lower value represents a less entangled (better) editing result.

\noindent\textbf{Learned Perceptual Image Patch Similarity (LPIPS).} LPIPS~\cite{zhang2018unreasonable} measures the diversity of a latent space. A larger LPIPS score indicates a more diverse space.

\subsection{Latent Space Interpolation and Editing}
\label{self_evaluation}
Our dual latent spaces achieve a certain degree of semantic separation, with $\mathcal{P}$-space controlling structural information like pose and $\mathcal{Z}$-space controlling texture information.

\noindent \textbf{Disentangled and Balanced Dual Spaces.} The two latent spaces in our architecture are both semantically meaningful while achieving desirable disentanglement. Specifically, the head pose is entirely controlled by the $\mathcal{P}$-space. As shown in \cref{fig:swap-celeb-z}, when re-sampling the $\bm{z}$ code, all the generated images share the same head pose. On the other hand, when $\bm{z}$ code is fixed, a similar texture, \ie, color, makeup, race, will appear on all results (\cref{fig:swap-celeb-p}). Besides, under the dual space setting, it is often desirable to have more balanced spaces rather than being dominant by a single space.
DAT~\cite{kwon2021diagonal} uses a diversity loss to encourage the diversity of its content space. In \cref{tab:lpips}, 
our dual spaces achieve a more balanced space separation than DAT while obtaining a higher overall diversity. The diversity difference between our dual spaces (\(\Delta_{LPIPS}\)) is $0.0662$, which is half of that of DAT~\cite{kwon2021diagonal}. This is consistent with our qualitative observation that both spaces of 
TransEditor shows higher controllability since they are more balanced.

\begin{figure}[]
  \centering
    \vspace{-0.2cm}
    \includegraphics[width=0.45\textwidth]{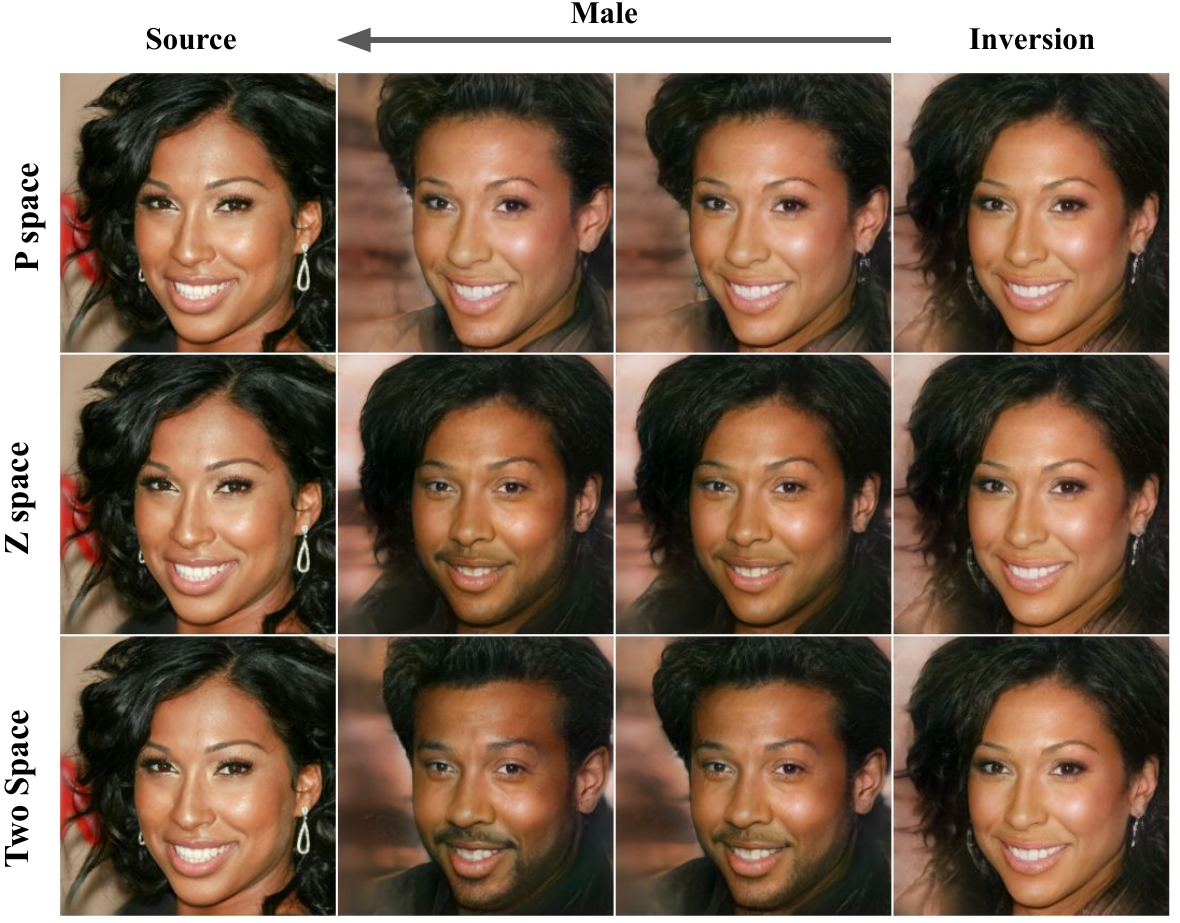}
    \vspace{-0.2cm}
    \caption{\textbf{Dual-Space Editing of attribute Male.} When editing using the $\bm{p}$ code (first row), the hair volume will decrease. And the face will gradually grow a small beard if using the $\bm{z}$ code (second row). The third row shows the result of editing the male attribute jointly via both spaces.}
   \label{fig:male-edit-two-space}
   \vspace{-0.35cm}
 \end{figure}

\begin{table}
  \centering
  \caption{The \textbf{LPIPS} score comparision on FFHQ-256~\cite{karras2019style}. A higher \textbf{LPIPS} represents a more diverse result. $\textbf{LPIPS}_{z}$ is obtained by fixing $\bm{p}$ code and randomly sampling $\bm{z}$ code, similar for $\textbf{LPIPS}_{p}$. \(\Delta_{LPIPS}\) is the difference between $\textbf{LPIPS}_{z}$ and $\textbf{LPIPS}_{p}$, and a lower value shows more balanced dual spaces.}
  \vspace{-0.2cm}
  \scalebox{0.85}{
  \begin{tabular}{c|c|c|c|c}
    \toprule
    Method & $\textbf{LPIPS}_{all}\uparrow$& $\textbf{LPIPS}_{z}$ & $\textbf{LPIPS}_{p} $ & \textbf{\(\Delta_{LPIPS}\)$\downarrow$} \\
    \midrule
    DAT~\cite{kwon2021diagonal} & 0.5596 & 0.5221 & 0.4071 & 0.1150 \\
    \midrule
    Ours & \textbf{0.5618} & 0.5118 & 0.4456 & \textbf{0.0662}\\
    \bottomrule
  \end{tabular}}
  \label{tab:lpips}
  \vspace{-0.3cm}
\end{table}

\begin{figure*}[]
    \vspace{-0.15cm}
    \begin{subfigure}[b]{0.49\textwidth}
        \includegraphics[width=\linewidth]{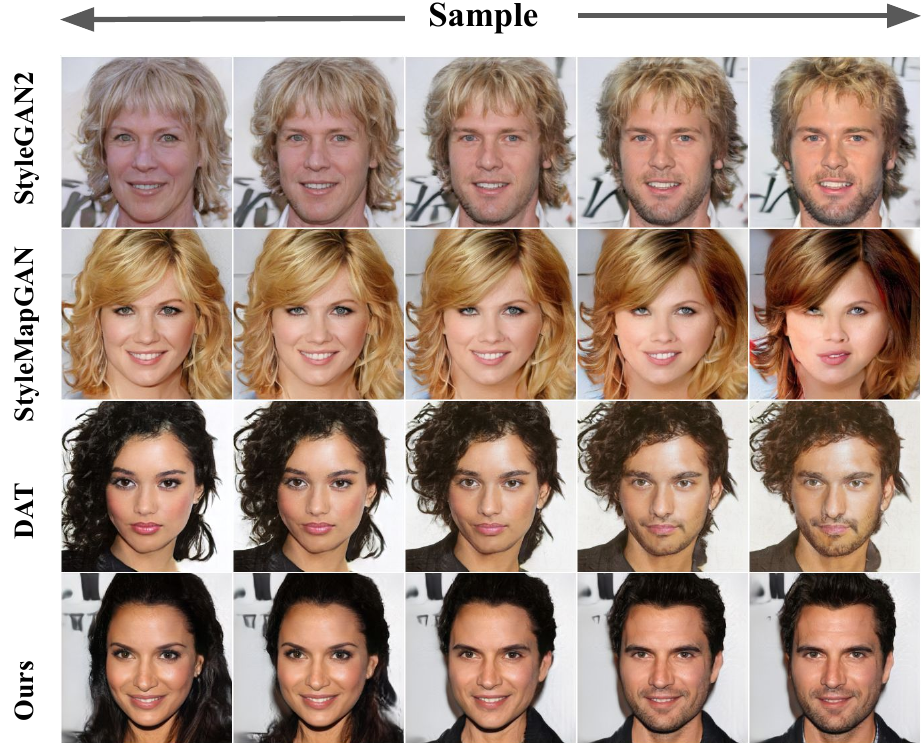}
            \caption{Gender}
            \label{fig:sample image editing_gender}
    \end{subfigure}%
        \hfill
    \begin{subfigure}[b]{0.49\textwidth}
        \includegraphics[width=\linewidth]{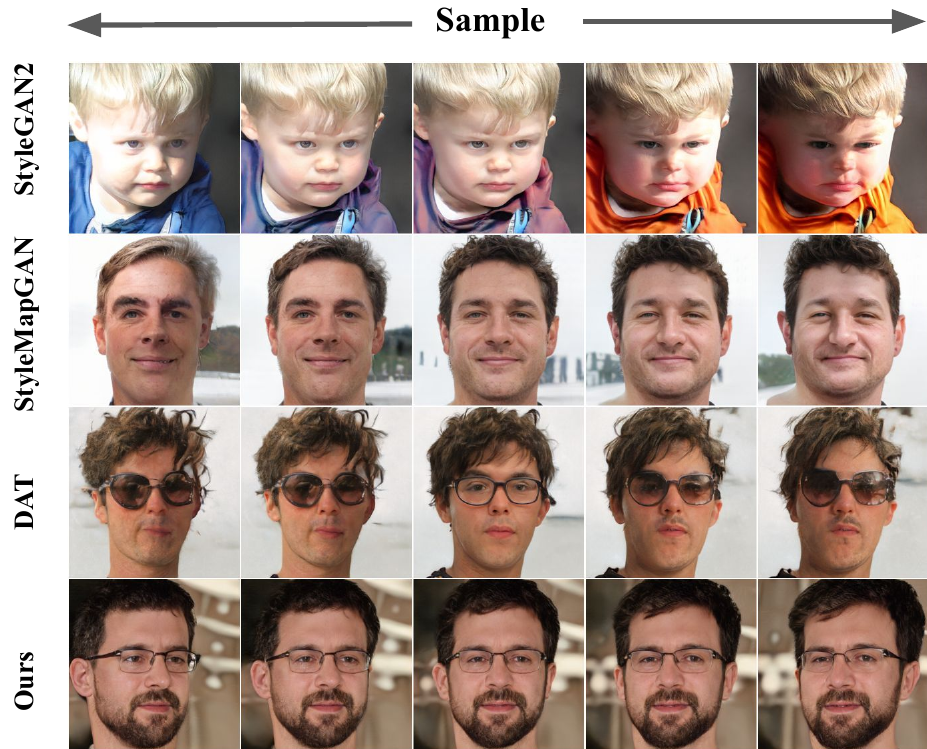}
            \caption{Pose}
            \label{fig:sample image editing_pose}
    \end{subfigure}%
    \vspace{-0.25cm}
    \caption{\textbf{Sample Image Editing Comparison}. Images on the third column are the sampled source images, which are semantically interpolated to the left and right sides. Our method preserves the identity better than other methods when editing gender (a) and pose (b).}
     \label{fig:sample image editing}
    \vspace{-0.3cm}
\end{figure*}

\noindent \textbf{Latent Space Interpolation.}  The two spaces are both smooth and semantic-aware, thus it is possible to change certain face attributes by interpolating towards a specific direction. \cref{fig:interp-celeb-z} shows a fixed head pose at each row and a smooth change in texture when $\bm{z}$ code is being interpolated. Similarly, \cref{fig:interp-celeb-p} shows small changes in texture information at each row with fixed $\bm{z}$ code, and the consistency in the head pose of each column with the same $\bm{p}$ code. The smoothness and semantic-aware property of the latent space allows better attribute editing.

\noindent \textbf{Dual Latent Space Editing.} Using the dual-space latent manipulation method (\cref{dual-space-editing}), attribute editing can be performed through linear interpolation towards a normal direction of the trained hyper-planes. The result in \cref{fig:male-edit-two-space} indicates that $p$ code controls the structural information such as hair volume, and $z$ code controls the texture information like the beard, and the joint editing results in the third row demonstrate the cooperation between the two latent spaces can achieve better editing performance. More editing results in~\cref{fig:teaser} further demonstrate the flexibility of our dual-space editing strategy.

\subsection{Comparison to State-of-the-art Methods}
\label{comparion_to_sota}

\subsubsection{Qualitative Evaluation}\label{qualitative}

\noindent \textbf{Sampled Image Editing.}
\cref{fig:sample image editing} shows the editing results of some sampled images. Since the change in gender may involve variation in both texture and structure information, the editing of gender is accomplished by editing on both $\mathcal{Z}$-space and $\mathcal{P}$-space (Content and Style space for DAT~\cite{kwon2021diagonal}).  For pose editing, only content space of DAT~\cite{kwon2021diagonal} and $\mathcal{P}$-space of ours are used. Compared with other methods, our method achieves better editing results. 
As mentioned in \cref{Related Work}, when the editing involves the style space of DAT~\cite{kwon2021diagonal}, the image tone could be easily changed (the third row of \cref{fig:sample image editing_gender}). The manipulation of the head pose is even a more challenging task, as the texture and structure information need to stay aligned to preserve the identity. \cref{fig:sample image editing_pose} shows the advantage of TransEditor. During pose editing, the $\bm{p}$ code is being manipulated when fixing the $\bm{z}$ code. However, the interaction process ensures that the style modulation parameter received by the generator has been aligned with the $\bm{p}$ code (\cref{interation formula}), thus producing a consistent texture throughout manipulation.

\begin{figure*}[]
   \begin{subfigure}[b]{0.49\textwidth}
        \includegraphics[width=\linewidth]{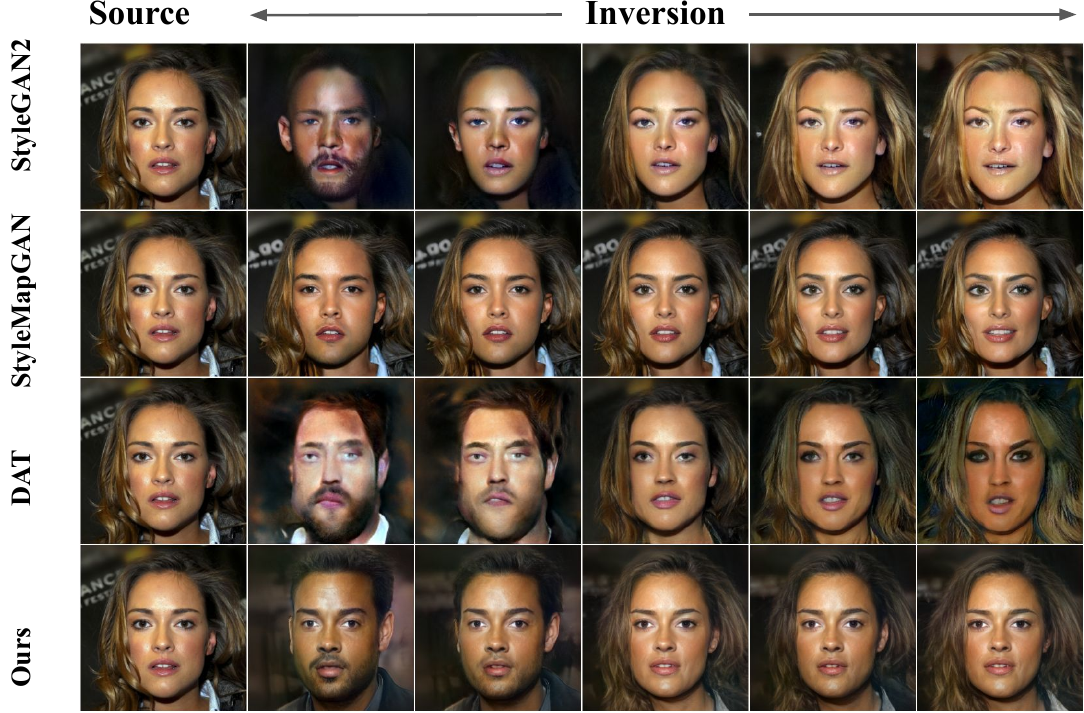}
            \caption{Gender}
            \label{fig:real_editing_gender}
        \end{subfigure}%
        \hfill
        \begin{subfigure}[b]{0.49\textwidth}
        \includegraphics[width=\linewidth]{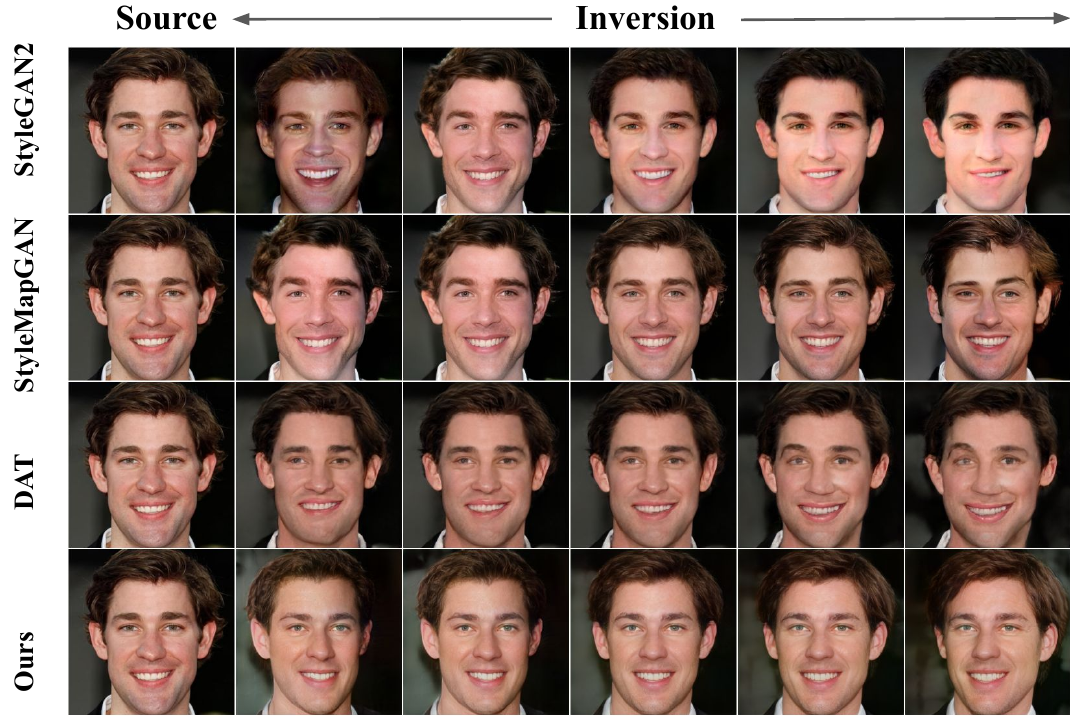}
            \caption{Pose}
            \label{fig:real_editing_pose}
        \end{subfigure}%
   \vspace{-0.16cm}
   \caption{\textbf{Real Image Editing Comparison}. Images on the first column are the real source images. The fourth column shows the reconstruction results, which are semantically interpolated to the left and right sides. Our method preserves the facial identity better than other methods when editing gender (a) and pose (b).} 
   \label{fig:real image editing}
\end{figure*}

\noindent \textbf{Real Image Editing.} 
To enable real image editing, we use the dual-space inversion method mentioned in \cref{dual-space-editing}. The latent manipulation method is identical to the sampled image editing. \cref{fig:real image editing} shows the comparative results on real image editing with state-of-the-art methods. In \cref{fig:real_editing_gender}, StyleMapGAN~\cite{kim2021exploiting} suffers from the global semantic editing, in which the face turns to male with long hair. The distortion of DAT~\cite{kwon2021diagonal} and attribute entanglement of StyleGAN2~\cite{karras2020analyzing} can be clearly observed. As shown in \cref{fig:real_editing_pose}, all the baselines fail to edit the head pose while our method obtains plausible results.

\subsubsection{Quantitative Evaluation} \label{quantitative}




For editing performance comparison, we chose three attributes: smile, head pose, and gender, which represent editing in $\mathcal{Z}$-space, $\mathcal{P}$-space, and both spaces, respectively. We use them to calculate our re-scoring metric. The comparative results of our method with StyleGAN2~\cite{karras2020analyzing}, StyleMapGAN~\cite{kim2021exploiting}, and DAT~\cite{kwon2021diagonal} are presented in \cref{tab: Re-scoring Comparision}. We observe that when editing each specific attribute, the influence of our model on other attributes is minimal, indicating that our method is the least entangled when performing editing. 

\begin{table*}
  \centering
  \caption{\textbf{Quantitative editing comparison between StyleGAN2~\cite{karras2020analyzing}, StyleMapGAN~\cite{kim2021exploiting}, DAT~\cite{kwon2021diagonal} and Ours.} The row-column entry represents the degree of change of column attribute while editing the row attribute. For example, the pose row and gender column shows the change of gender when editing the head pose. The comparison shows our method has the least effect on other attributes during editing.}
  \vspace{-0.2cm}
  \scalebox{0.85}{
  \begin{tabular}{c|c c c c |c c c c |c c c c }
    \toprule
     & &\textbf{Pose}$\downarrow$& & & &\textbf{Gender}$\downarrow$& &  & &\textbf{Smile}$\downarrow$& &\\
    \midrule
     
     Method& StyleGAN2& StyleMap& DAT& Ours &StyleGAN2 & StyleMap&DAT& Ours&StyleGAN2& StyleMap&DAT& Ours \\
    \midrule 
    \textbf{Pose} &-&-&-&- &0.683&3.418&4.869&\textbf{0.231} &0.577&0.947&0.625&\textbf{0.153}\\
    \textbf{Gender} &0.738&0.340&0.386&\textbf{0.313}&-&-&-&- &0.189&1.804&0.514&\textbf{0.035}\\
    
    \textbf{Smile} &0.0712&0.181&0.034&\textbf{0.031} &0.146&0.146&0.111&\textbf{0.040} &-&-&-&- \\
    
    \bottomrule
  \end{tabular}}
  \label{tab: Re-scoring Comparision}
\end{table*}

\subsection{Ablation Study}
\label{ablation_study}

%

\begin{figure}[]
  \centering
    \includegraphics[width=0.46\textwidth]{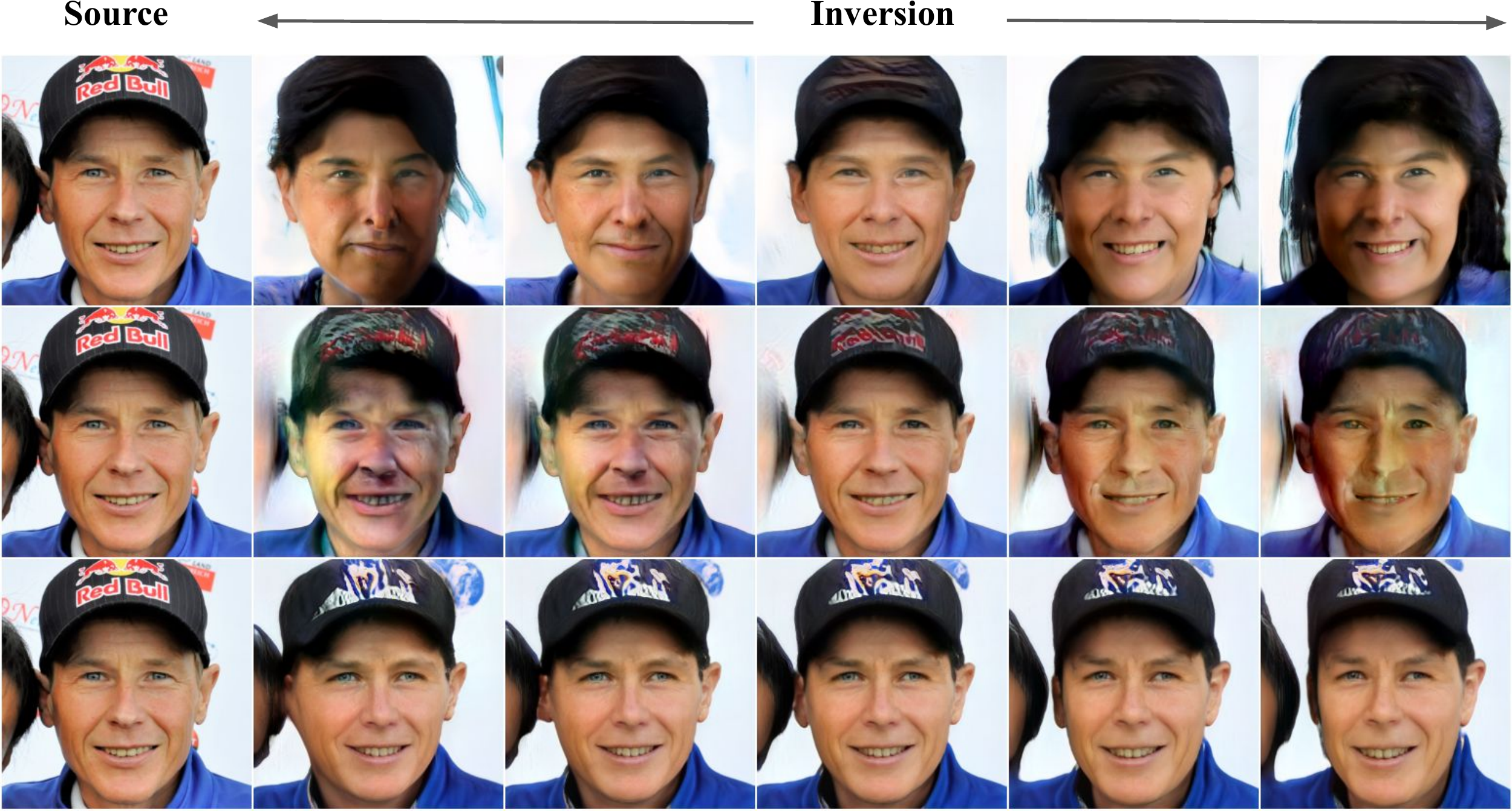}
    \vspace{-0.1cm}
    \caption{\textbf{Ablation Study on the Transformer and the dual-space design}. Rows from top to bottom showcase editing results of head pose using TransEditor w/o Transformer, TransEditor w/o dual spaces, and TransEditor itself, respectively.}
  \label{fig:ablation-edit}
  \vspace{-0.2cm}
 \end{figure}

\begin{figure}
  \centering
    \begin{subfigure}{0.23\textwidth}
    \includegraphics[width=\textwidth]{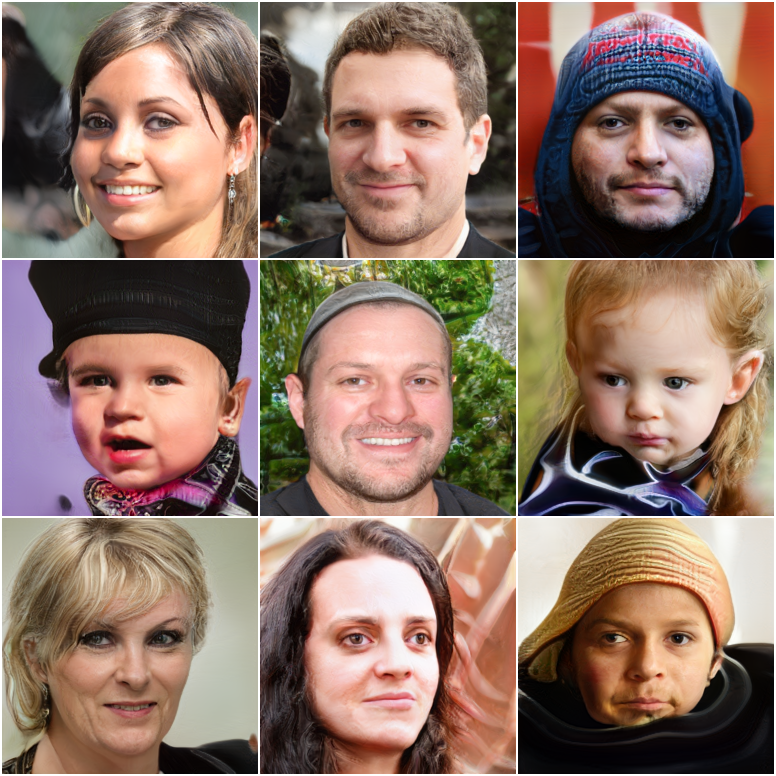} 
    \caption{Z as Q matrix, P as K,V matrix}
    \label{fig:z-as-q p-k-v}
  \end{subfigure}
  \hfill
  \begin{subfigure}{0.23\textwidth}
    \includegraphics[width=\textwidth]{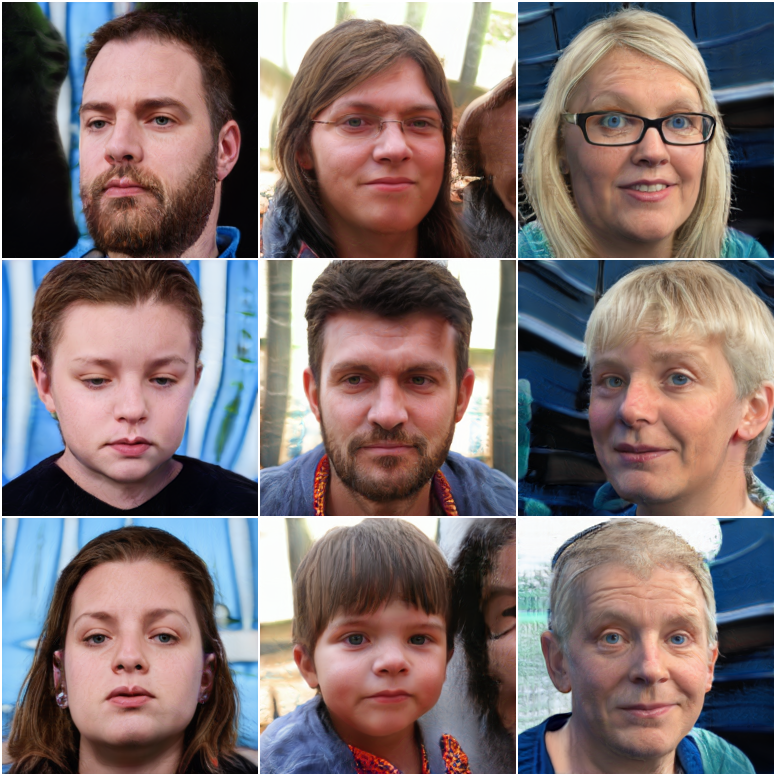}
    \caption{P as Q matrix, Z as K,V matrix}
    \label{fig:p-as-q z-k-v}
  \end{subfigure}
  \label{fig:ablation-kqv}
  \vspace{-0.1cm}
  \caption{\textbf{Ablation study on different interaction methods using Transformer.} Both images are generated by fixing $\bm{z}$ code and sampling $\bm{p}$ code at each column. Figure(b) is our current setting, which shows a more disentangled result than figure(a).}

  \vspace{-0.2cm}
  \end{figure}

\noindent
\textbf{Space Interaction via Transformer.}
The space interaction is crucial in our architecture for the semantic balance between the two spaces. Removing the interaction process (\ie, the generator receives two completely independent codes) 
produces an unbalanced result. Under this setting, the generated images are mostly controlled by the $\mathcal{P}$-space, \ie, the style information such as hue is also contained in the $\mathcal{P}$-space. Since the $\mathcal{P}$-space controls the majority of information, it is more entangled during editing, leading to joint changes in color and shape (the first row of~\cref{fig:ablation-edit}). In contrast, with the cross-space interaction mechanism, the tones of the images keep consistent in a successful pose editing (the third row of~\cref{fig:ablation-edit}). Therefore, establishing a connection between the two spaces enables a more balanced setting, which benefits facial attribute editing. We include further re-scoring quantitative evaluation in \cref{tab: Ablation re-scoring Comparison}. Our method clearly outperforms the one w/o Transformer.

\begin{table}
  \centering
  \caption{\textbf{Re-scoring metrics of w/o or w/ Transformer}.}
  \vspace{-0.2cm}
  \scalebox{0.8}{
  \begin{tabular}{c|c c |c c}
    \toprule
     & \textbf{Pose}$\downarrow$&  &\textbf{Gender}$\downarrow$ \\
    \midrule
     
     Method& w/o Transformer & Ours & w/o Transformer & Ours \\
    \midrule 
    \textbf{Pose} &-&-  &0.504&\textbf{0.231} \\
    \textbf{Gender} &0.782& \textbf{0.313}&-&- \\
    
    \bottomrule
  \end{tabular}}
  \label{tab: Ablation re-scoring Comparison}
  \vspace{-0.5cm}
\end{table}


\noindent
\textbf{Dual-Space Design v.s. Single-Space Design.}
We then evaluate the role of the dual-space design. TransStyleGAN~\cite{vaswani2017attention} employs the self-attention mechanism to establish connections between different style codes within the single-space design. However, as can be seen in the second row of ~\cref{fig:ablation-edit}, when the step size of pose-editing gets larger, the face orientation is still difficult to change, while the distortion of the face and the change of tone could be serious. This indicates the difficulty of editing pose in a single space, demonstrating the advantage of our dual latent spaces.

\noindent
\textbf{Selection of K, Q, V Matrices.}\label{ablation_kqv}
For the Cross-space Interaction module, we need to consider the choice of the $\mathbf{K}, \mathbf{Q}, \mathbf{V}$ matrices for the cross-space attention. Some studies in the multi-modality field use the single-modality feature that needs to be refined as the query, and the other modality feature as the key and value. 
In our case, this is equivalent to using the $\bm{z}^{+}$ space as the query matrix and the $\bm{p}^{+}$ space as the key and value matrix. However, the output of the cross-space attention module, as the refined feature of the $\bm{z}^{+}$ space, is a weighting of the value matrix ($\bm{p}^{+}$ space), which may produce certain entanglement of the two spaces. This setting (\cref{fig:z-as-q p-k-v}) shows a severe entanglement between $\mathcal{P}$-space and $\mathcal{Z}$-space that swapping $\bm{p}$ codes result in different textures, which is not desirable for editing. On the other hand, our setting shown in \cref{fig:p-as-q z-k-v} is much better disentangled, since the $\bm{p}^{+}$ space is only adopted to as query to help update the $\bm{z}^{+}$ space. Therefore, this interaction method in \cref{Transformer-based Cross-space Interaction} is more desirable for our design. 
\section{Conclusion and Discussion} 
This paper introduces TransEditor, a novel dual-space GAN architecture with a Cross-Space Interaction mechanism based on the Transformer. Besides, we propose a new dual-space image editing and inversion strategy for highly controllable facial editing.
Extensive experiments show the effectiveness of TransEditor in attribute disentanglement and controllability, surpassing state-of-the-art baselines in complicated attribute editing.
The proposed TransEditor is readily applicable to many real-world applications such as photo retouching and face manipulation, which however, might be used unethically. Devising better media forensics approaches could be countermeasures.
As for limitations, the editing process relies on auxiliary classifiers (for semantic boundaries), whose quality and diversity may limit editable attributes. In addition, the improvement on the cross-space interaction of a dual-space GAN for editing tasks can be interesting future work.

\noindent
\textbf{Acknowledgement.} This study is partly supported under the RIE2020 Industry Alignment Fund Industry Collaboration Projects (IAF-ICP) Funding Initiative, as well as cash and in-kind contribution from the industry partner(s).


{\small
\bibliographystyle{ieee_fullname}
\bibliography{arxiv}

\begin{thebibliography}{10}\itemsep=-1pt

\bibitem{abdal2019image2stylegan}
Rameen Abdal, Yipeng Qin, and Peter Wonka.
\newblock Image2stylegan: How to embed images into the stylegan latent space?
\newblock In {\em ICCV}, 2019.

\bibitem{alaluf2021restyle}
Yuval Alaluf, Or Patashnik, and Daniel Cohen-Or.
\newblock Restyle: A residual-based stylegan encoder via iterative refinement.
\newblock In {\em ICCV}, 2021.

\bibitem{albahar2021pose}
Badour Albahar, Jingwan Lu, Jimei Yang, Zhixin Shu, Eli Shechtman, and Jia-Bin
  Huang.
\newblock Pose with style: Detail-preserving pose-guided image synthesis with
  conditional stylegan.
\newblock {\em TOG}, 40, 2021.

\bibitem{alharbi2020disentangled}
Yazeed Alharbi and Peter Wonka.
\newblock Disentangled image generation through structured noise injection.
\newblock In {\em CVPR}, 2020.

\bibitem{carion2020end}
Nicolas Carion, Francisco Massa, Gabriel Synnaeve, Nicolas Usunier, Alexander
  Kirillov, and Sergey Zagoruyko.
\newblock End-to-end object detection with transformers.
\newblock In {\em ECCV}, 2020.

\bibitem{chai2021ensembling}
Lucy Chai, Jun-Yan Zhu, Eli Shechtman, Phillip Isola, and Richard Zhang.
\newblock Ensembling with deep generative views.
\newblock In {\em CVPR}, 2021.

\bibitem{collins2020editing}
Edo Collins, Raja Bala, Bob Price, and Sabine Susstrunk.
\newblock Editing in style: Uncovering the local semantics of gans.
\newblock In {\em CVPR}, 2020.

\bibitem{cornia2020meshed}
Marcella Cornia, Matteo Stefanini, Lorenzo Baraldi, and Rita Cucchiara.
\newblock Meshed-memory transformer for image captioning.
\newblock In {\em CVPR}, 2020.

\bibitem{ding2021cogview}
Ming Ding, Zhuoyi Yang, Wenyi Hong, Wendi Zheng, Chang Zhou, Da Yin, Junyang
  Lin, Xu Zou, Zhou Shao, Hongxia Yang, et~al.
\newblock Cogview: Mastering text-to-image generation via transformers.
\newblock {\em arXiv preprint arXiv:2105.13290}, 2021.

\bibitem{dosovitskiy2021an}
Alexey Dosovitskiy, Lucas Beyer, Alexander Kolesnikov, Dirk Weissenborn,
  Xiaohua Zhai, Thomas Unterthiner, Mostafa Dehghani, Matthias Minderer, Georg
  Heigold, Sylvain Gelly, Jakob Uszkoreit, and Neil Houlsby.
\newblock An image is worth 16x16 words: Transformers for image recognition at
  scale.
\newblock In {\em ICLR}, 2021.

\bibitem{gardner1998artificial}
Matt~W Gardner and SR Dorling.
\newblock Artificial neural networks (the multilayer perceptron)—a review of
  applications in the atmospheric sciences.
\newblock {\em Atmospheric environment}, 32:2627--2636, 1998.

\bibitem{ghiasi2017exploring}
Golnaz Ghiasi, Honglak Lee, Manjunath Kudlur, Vincent Dumoulin, and Jonathon
  Shlens.
\newblock Exploring the structure of a real-time, arbitrary neural artistic
  stylization network.
\newblock {\em arXiv preprint arXiv:1705.06830}, 2017.

\bibitem{goetschalckx2019ganalyze}
Lore Goetschalckx, Alex Andonian, Aude Oliva, and Phillip Isola.
\newblock Ganalyze: Toward visual definitions of cognitive image properties.
\newblock In {\em ICCV}, 2019.

\bibitem{goodfellow2014generative}
Ian Goodfellow, Jean Pouget-Abadie, Mehdi Mirza, Bing Xu, David Warde-Farley,
  Sherjil Ozair, Aaron Courville, and Yoshua Bengio.
\newblock Generative adversarial nets.
\newblock {\em NeurIPS}, 27, 2014.

\bibitem{gu2020image}
Jinjin Gu, Yujun Shen, and Bolei Zhou.
\newblock Image processing using multi-code gan prior.
\newblock In {\em CVPR}, 2020.

\bibitem{harkonen2020ganspace}
Erik H{\"a}rk{\"o}nen, Aaron Hertzmann, Jaakko Lehtinen, and Sylvain Paris.
\newblock Ganspace: Discovering interpretable gan controls.
\newblock {\em NeurIPS}, 2020.

\bibitem{he2019attgan}
Zhenliang He, Wangmeng Zuo, Meina Kan, Shiguang Shan, and Xilin Chen.
\newblock Attgan: Facial attribute editing by only changing what you want.
\newblock {\em TIP}, 28:5464--5478, 2019.

\bibitem{heusel2017gans}
Martin Heusel, Hubert Ramsauer, Thomas Unterthiner, Bernhard Nessler, and Sepp
  Hochreiter.
\newblock Gans trained by a two time-scale update rule converge to a local nash
  equilibrium.
\newblock {\em NeurIPS}, 2017.

\bibitem{huang2019attention}
Lun Huang, Wenmin Wang, Jie Chen, and Xiao-Yong Wei.
\newblock Attention on attention for image captioning.
\newblock In {\em ICCV}, 2019.

\bibitem{huang2017arbitrary}
Xun Huang and Serge Belongie.
\newblock Arbitrary style transfer in real-time with adaptive instance
  normalization.
\newblock In {\em ICCV}, 2017.

\bibitem{gansteerability}
Ali Jahanian, Lucy Chai, and Phillip Isola.
\newblock On the "steerability" of generative adversarial networks.
\newblock In {\em ICLR}, 2020.

\bibitem{jiang2021transgan}
Yifan Jiang, Shiyu Chang, and Zhangyang Wang.
\newblock Transgan: Two transformers can make one strong gan.
\newblock {\em arXiv preprint arXiv:2102.07074}, 2021.

\bibitem{karras2017progressive}
Tero Karras, Timo Aila, Samuli Laine, and Jaakko Lehtinen.
\newblock Progressive growing of gans for improved quality, stability, and
  variation.
\newblock {\em arXiv preprint arXiv:1710.10196}, 2017.

\bibitem{karras2019style}
Tero Karras, Samuli Laine, and Timo Aila.
\newblock A style-based generator architecture for generative adversarial
  networks.
\newblock In {\em CVPR}, 2019.

\bibitem{karras2020analyzing}
Tero Karras, Samuli Laine, Miika Aittala, Janne Hellsten, Jaakko Lehtinen, and
  Timo Aila.
\newblock Analyzing and improving the image quality of stylegan.
\newblock In {\em CVPR}, 2020.

\bibitem{kim2021exploiting}
Hyunsu Kim, Yunjey Choi, Junho Kim, Sungjoo Yoo, and Youngjung Uh.
\newblock Exploiting spatial dimensions of latent in gan for real-time image
  editing.
\newblock In {\em CVPR}, 2021.

\bibitem{kwon2021diagonal}
Gihyun Kwon and Jong~Chul Ye.
\newblock Diagonal attention and style-based gan for content-style
  disentanglement in image generation and translation.
\newblock In {\em ICCV}, 2021.

\bibitem{lewis2021tryongan}
Kathleen~M Lewis, Srivatsan Varadharajan, and Ira Kemelmacher-Shlizerman.
\newblock Tryongan: Body-aware try-on via layered interpolation.
\newblock {\em TOG}, 40, 2021.

\bibitem{li2021transforming}
Heyi Li, Jinlong Liu, Yunzhi Bai, Huayan Wang, and Klaus Mueller.
\newblock Transforming the latent space of stylegan for real face editing.
\newblock {\em arXiv preprint arXiv:2105.14230}, 2021.

\bibitem{ma2019invertibility}
Fangchang Ma, Ulas Ayaz, and Sertac Karaman.
\newblock Invertibility of convolutional generative networks from partial
  measurements.
\newblock {\em NeurIPS}, 2019.

\bibitem{park2020swapping}
Taesung Park, Jun-Yan Zhu, Oliver Wang, Jingwan Lu, Eli Shechtman, Alexei
  Efros, and Richard Zhang.
\newblock Swapping autoencoder for deep image manipulation.
\newblock {\em NeurIPS}, 33, 2020.

\bibitem{ramesh2021zero}
Aditya Ramesh, Mikhail Pavlov, Gabriel Goh, Scott Gray, Chelsea Voss, Alec
  Radford, Mark Chen, and Ilya Sutskever.
\newblock Zero-shot text-to-image generation.
\newblock {\em arXiv preprint arXiv:2102.12092}, 2021.

\bibitem{richardson2021encoding}
Elad Richardson, Yuval Alaluf, Or Patashnik, Yotam Nitzan, Yaniv Azar, Stav
  Shapiro, and Daniel Cohen-Or.
\newblock Encoding in style: a stylegan encoder for image-to-image translation.
\newblock In {\em CVPR}, 2021.

\bibitem{Shen_2020_CVPR}
Yujun Shen, Jinjin Gu, Xiaoou Tang, and Bolei Zhou.
\newblock Interpreting the latent space of gans for semantic face editing.
\newblock In {\em CVPR}, 2020.

\bibitem{shen2020interfacegan}
Yujun Shen, Ceyuan Yang, Xiaoou Tang, and Bolei Zhou.
\newblock Interfacegan: Interpreting the disentangled face representation
  learned by gans.
\newblock {\em PAMI}, 2020.

\bibitem{shen2021closed}
Yujun Shen and Bolei Zhou.
\newblock Closed-form factorization of latent semantics in gans.
\newblock In {\em CVPR}, 2021.

\bibitem{spingarn2020gan}
Nurit Spingarn, Ron Banner, and Tomer Michaeli.
\newblock Gan" steerability" without optimization.
\newblock In {\em ICLR}, 2020.

\bibitem{szegedy2016rethinking}
Christian Szegedy, Vincent Vanhoucke, Sergey Ioffe, Jon Shlens, and Zbigniew
  Wojna.
\newblock Rethinking the inception architecture for computer vision.
\newblock In {\em CVPR}, 2016.

\bibitem{tov2021designing}
Omer Tov, Yuval Alaluf, Yotam Nitzan, Or Patashnik, and Daniel Cohen-Or.
\newblock Designing an encoder for stylegan image manipulation.
\newblock {\em ACM TOG}, 40:1--14, 2021.

\bibitem{vaswani2017attention}
Ashish Vaswani, Noam Shazeer, Niki Parmar, Jakob Uszkoreit, Llion Jones,
  Aidan~N Gomez, {\L}ukasz Kaiser, and Illia Polosukhin.
\newblock Attention is all you need.
\newblock In {\em NeurIPS}, 2017.

\bibitem{wang2021end}
Yuqing Wang, Zhaoliang Xu, Xinlong Wang, Chunhua Shen, Baoshan Cheng, Hao Shen,
  and Huaxia Xia.
\newblock End-to-end video instance segmentation with transformers.
\newblock In {\em CVPR}, 2021.

\bibitem{wu2021stylespace}
Zongze Wu, Dani Lischinski, and Eli Shechtman.
\newblock Stylespace analysis: Disentangled controls for stylegan image
  generation.
\newblock In {\em CVPR}, 2021.

\bibitem{zhang2018generative}
Gang Zhang, Meina Kan, Shiguang Shan, and Xilin Chen.
\newblock Generative adversarial network with spatial attention for face
  attribute editing.
\newblock In {\em ECCV}, 2018.

\bibitem{zhang2018unreasonable}
Richard Zhang, Phillip Isola, Alexei~A Efros, Eli Shechtman, and Oliver Wang.
\newblock The unreasonable effectiveness of deep features as a perceptual
  metric.
\newblock In {\em CVPR}, 2018.

\bibitem{zhu2020domain}
Jiapeng Zhu, Yujun Shen, Deli Zhao, and Bolei Zhou.
\newblock In-domain gan inversion for real image editing.
\newblock In {\em ECCV}, 2020.

\bibitem{zhu2021barbershop}
Peihao Zhu, Rameen Abdal, John Femiani, and Peter Wonka.
\newblock Barbershop: Gan-based image compositing using segmentation masks.
\newblock {\em TOG}, 2021.

\bibitem{zhu2020improved}
Peihao Zhu, Rameen Abdal, Yipeng Qin, John Femiani, and Peter Wonka.
\newblock Improved stylegan embedding: Where are the good latents?
\newblock {\em arXiv preprint arXiv:2012.09036}, 2020.

\end{thebibliography}
}

\clearpage

\setcounter{section}{0}
\renewcommand\thesection{\Alph{section}}

\section{Detailed Calculation of Metrics}

\noindent\textbf{Re-scoring Calculation.}
This designed metric is used to quantitatively evaluate the editing performance. It is desirable that when editing, the edited attribute will change towards the targeting direction as much as possible, while other attributes remain as less impacted as possible. For example, when editing an attribute towards the plus direction, we expect the score to increase. The amount of change could be quantitatively evaluated using trained classifiers. More specifically, by adding the score difference of the edited attribute between each editing step, the accumulated change could be calculated. Denote the accumulated change of edited attribute as \(C_{e}\), that of the influenced attribute as \(C_{i}\), then it is optimal when \(C_e\) is as large as possible and \(C_i\) is as small as possible. Therefore, the ratio \(C_i/C_{e}\) measures the degree of another attribute being influenced when performing the editing. Note if only \(C_{i}\) is measured as evaluation, the value will be the smallest between two identical images, thus failing to describe the editing performance. Moreover, when the value of \(C_i\) are identical, a larger \(C_{e}\) represents more change in the desired attribute, which is desirable for the editing task.   

In our experiments, for each attribute, we generate 4,000 images and perform editing. Then the corresponding trained attribute classifier~\cite{chai2021ensembling} is used to re-score the edited images, resulting in 28,000 scores (6 steps and the origin image) for each attribute. 

\noindent\textbf{Identity Re-scoring Calculation.}
To qualitatively evaluate the change of identity during editing, we also utilize the trained Inception v3~\cite{szegedy2016rethinking} model to extract perceptual features from images. The calculation of this metric is similar to the Re-scoring Calculation above. The cosine similarity of the extracted feature between the images at each step will be calculated, and its accumulated value, denoted as \(C_{id}\), measures the amount of change in identity. The meaning and calculation of \(C_{e}\) are identical to the Re-scoring Calculation. A smaller value of \(C_{id}/C_{e}\) means better preservation of identity when editing the attribute. 


\noindent\textbf{Learned Perceptual Image Patch Similarity (LPIPS).} LPIPS~\cite{zhang2018unreasonable} measures the diversity of a latent space. A larger LPIPS score indicates a more diverse space. Since there are two spaces, we perform this calculation similar to DAT~\cite{kwon2021diagonal}. $\mathbf{LPIPS}_{z}$ is calculated by sampling 40 $\bm{z}$ codes with a fixed $\bm{p}$ code. Similarly, $\mathbf{LPIPS}_{p}$ is calculated by sampling 40 $\bm{z}$ codes with a fixed $\bm{p}$ code. For $\mathbf{LPIPS}_{all}$, it is calculated by sampling 40 pairs of $\bm{z}$ and $\bm{p}$ codes. All the processes are repeated by 1,000 times. 

\noindent\textbf{Frechet inception distance (FID).} FID~\cite{heusel2017gans} measures the image generation quality by calculating the feature difference between the real images and the generated images. A smaller FID value implies a better generation quality.

\section{Datasets}

\noindent\textbf{CelebA-HQ.}  CelebA-HQ~\cite{karras2017progressive} contains 30,000 celebrity face images with a resolution of $1024 \times 1024$. The images are annotated with $40$ attribute labels. 

\noindent\textbf{FFHQ.} FFHQ~\cite{karras2019style} contains 70,000 high-quality face images with a resolution of $1024 \times 1024$.
FFHQ contains more changes in terms of hue, age, and background than CelebA-HQ.

\section{Implementation Details}

\subsection{Dual Latent Space and Mapping Functions}
In our experiments, both the dimension of $\mathcal{Z}$-space and $\mathcal{P}$-space are set to be $16\times 512$. Each latent vector $\bm{z}_{i} \in 1\times512$ and $\bm{p}_{i} \in 1\times512$. Their corresponding mapping functions, $M_{\bm{z}_{i}}$ and $M_{\bm{p}_{i}}$ are MLPs that map $\bm{z}_{i}$ to $\bm{z}^{+}_{i} \in 1\times512$ and $\bm{p}_{i}$ to $\bm{p}^{+}_{i} \in 1\times512$. 

\subsection{Transformer-Based Interaction}
For interaction at each layer, we utilize a Transformer-based multi-head attention module. In our model, we set the number of Transformer layers to be $8$ and the dimension of the latent code input to Transformer to be $512$. The number of heads in the multi-head cross attention module is set to be $8$ so the dimensionality for each head is $d_k = 512/8 = 64$. Besides, we add positional encoding to both latent codes from $\mathcal{Z}$-space and $\mathcal{P}$-space before the first Transformer layer. The positional encoding matrix is an identity matrix of size $16 \times 16$.

\subsection{Training Details}
Since two latent spaces are used in the proposed TransEditor, the optimization objective can be written as:
\begin{equation}
    \begin{aligned}
    \underset{\mathbf{G}}{min}\,\underset{\mathbf{D}}{max}\, V(\mathbf{D},\mathbf{G}) = E_{x\sim p_{data}(x)}[\log{\mathbf{D}(x)}]+ \\ E_{(\bm{z},\bm{p})\sim p_{(\mathcal{Z}\times \mathcal{P})}(\bm{z},\bm{p})}[\log(1-\mathbf{D}(\mathbf{G}(\bm{z},\bm{p})))].
    \end{aligned}
\end{equation}

As mentioned, we only apply the adversarial loss~\cite{goodfellow2014generative} and Path Length Regularization used in StyleGAN2~\cite{karras2020analyzing}. For the adversarial loss, similar to StyleGAN2, it is composed of non-saturating loss, \ie, \(f(t)=softplus(t) = log(1+exp(t))\). 
For the generator,

\begin{equation}
    L_{\mathbf{G}} = \lambda_{adv}f(-\mathbf{D}(\mathbf{G}(\bm{z},\bm{p}))) + \lambda_{path\_regu}L_{path\_regu},
\end{equation}
and $L_{path\_regu}$ is the path length regularization.
For the discriminator, 

\begin{equation}
\begin{aligned}
    L_{\mathbf{D}} = &\lambda_{dis}[f(\mathbf{D}(X_{fake})) +f(-\mathbf{D}(X_{real})) ]   \\ 
    &+ \lambda_{d\_regu}L_{d\_regu},
    \end{aligned}
\end{equation}
where $L_{d\_regu}$ is the gradient regularization for the discriminator.  

The training of FFHQ~\cite{karras2019style} and CelebA-HQ~\cite{karras2017progressive} are performed on the resolution of $256\times256$. We set $\lambda_{d\_regu}$ to $10$, $L_{path\_regu}$ to $2$, $\lambda_{adv}$ and $\lambda_{dis}$ to $1$ in our training. For CelebA-HQ, we utilize 29,000 images as the training set, with 1,000 left for testing. Then we train the model to 370,000 iterations with a batch size of 16 using 8 cards.
For FFHQ, we utilize 69,000 and 1,000 images for training and testing, respectively. The model is trained to 800,000 iterations with the batch size of $16$ on a single card. 

\subsection{Dual Space Inversion and Editing}
The loss functions used in our Dual Space Inversion are similar to pSp~\cite{richardson2021encoding}. We apply the same pixel-wise $\mathcal{L}_{2}$ loss, LPIPS loss, and ID loss as in pSp~\cite{richardson2021encoding}. Their weight is set to be $1.0$, $0.8$, and $0.1$, respectively. For both datasets, we train the Dual Space Inversion network to 500,000 iterations, with a batch size of $8$. 

For Dual Space Editing, auxiliary attribute classifies~\cite{chai2021ensembling} are used. Specifically, we randomly sample 150,000 pairs of $\bm{z} \in \mathbb{R}^{n\times512}$ codes and $\bm{p} \in \mathbb{R}^{n\times512}$ codes from a standard normal distribution and map them to $\bm{z}^{+}$ and $\bm{p}^{+}$ for image generation. Then for each attribute, the corresponding classifier will be used for scoring on the generated images. We then train an SVM classifier with $\bm{z}^{+}$ and the score for attribute $i$ as training data and labels, thus finding the normal vector $n_z$ of the partition interface corresponding to attribute $i$ in $\mathcal{Z}^{+}$-space. Similarly, we obtain the normal vector $n_p$ in $\mathcal{P}^{+}$-space. Thereafter, we can move $\lambda_{z}$ steps along $n_z$ and $\lambda_{p}$ steps along $n_p$ to get the new latent codes $(\bm{z}^{+} + \lambda_{z} * n_z, \bm{p}^{+} + \lambda_{p} * n_p)$. We can flexibly adjust $\lambda_z$ and $\lambda_p$ to control the contribution of each space to the final editing of different attributes. For example, if we set $\lambda_z$ to $0$, $\mathcal{Z}^{+}$-space is fixed, and only $\mathcal{P}^{+}$-space is altered. This diagram can be applied to smile editing (\cref{fig:edit_smile_celeba}). Similarly, if we set $\lambda_p$ to $0$, $\mathcal{P}^{+}$-space is fixed, and only $\mathcal{Z}^{+}$-space is altered. This diagram can be applied to pose editing (\cref{fig:edit_pose_celeba}, \cref{fig:edit_pose_ffhq}). For gender (\cref{fig:edit_gender_ffhq}) and age (\cref{fig:edit_age_ffhq}) editing, both $\lambda_{z}$ and $\lambda_{p}$ are adjusted.

\section{More Results and Analysis}

\subsection{More Ablation Study}


\noindent\textbf{Trials of Training Techniques.}
The Path Length Regularization in StyleGAN2~\cite{karras2020analyzing} is used to smooth the latent space $\mathcal{W}$, which is minimized when changing a fixed step of the latent code will result in a fixed-magnitude change in the image. In our design, the space that corresponds to $\mathcal{W}$ in StyleGAN2~\cite{karras2020analyzing} is the output of our cross-space interaction. Since it is not desired that any certain layer of the $\mathcal{W}$ space be dominant, we utilize the same regularization loss on our $\mathcal{W}$ space. 

We have also experimented to add the regularization loss on the $\mathcal{P^{+}}$-space. The result shows that this will discourage the $\mathcal{P^{+}}$ to influence the final result. When this regularization loss is added on the $\mathcal{P^{+}}$-space, changing the entire $\bm{p}$ code will only result in a little change in the generated image, which is undesirable since we need more balanced dual spaces for editing. 

\noindent\textbf{Number of Transformer.}
The number of Transformer layers is related to the degree of interaction. When the number of layers gets larger, the $\bm{z}^{+}$ code will be queried by the \(p^{+}\) code more, resulting in a stronger correlation. We experimented with different number of Transformer, and the result shows that using more layers will result in better head pose consistency when \(p^{+}\) code is fixed, although the difference is not significant. In most of our settings, we utilize $8$ 
layers of Transformers.

\noindent\textbf{Alternative interaction module (\eg, MLP).}
The ablation study on Space Interaction via Transformer has shown its cruciality.
Our design of using $\mathcal{P}$-space as the query establishes a connection between the two spaces while ensuring their disentanglement. We have also tried other interaction modules, \eg, MLP.
Although MLP can create a linkage, the results of which are inferior to the Transformer-based interaction module due to the entanglement caused by MLP. For instance, the Pose-Identity re-scoring~($\downarrow$) result of the MLP variant is $7.024$ compared with our result of $\bf2.326$


\subsection{More Comparison with State of the Art}

\begin{figure*}[]
    \vspace{-0.10cm}
    \begin{subfigure}[b]{0.49\textwidth}
        \includegraphics[width=\linewidth]{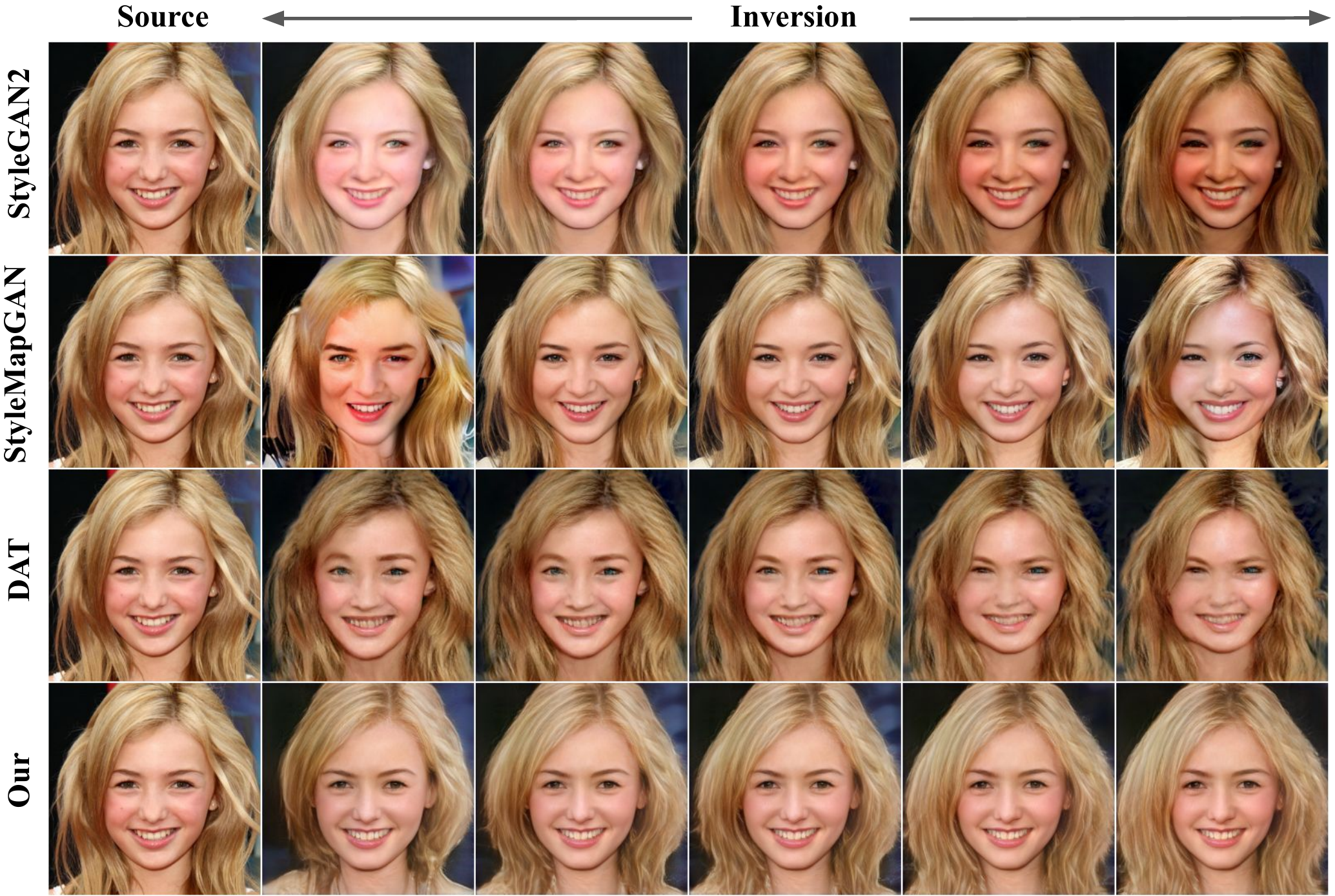}
            \caption{Wavy Hair}
            \label{fig:edit_real_wavy_hair}
        \end{subfigure}%
        \hfill
        \begin{subfigure}[b]{0.49\textwidth}
        \includegraphics[width=\linewidth]{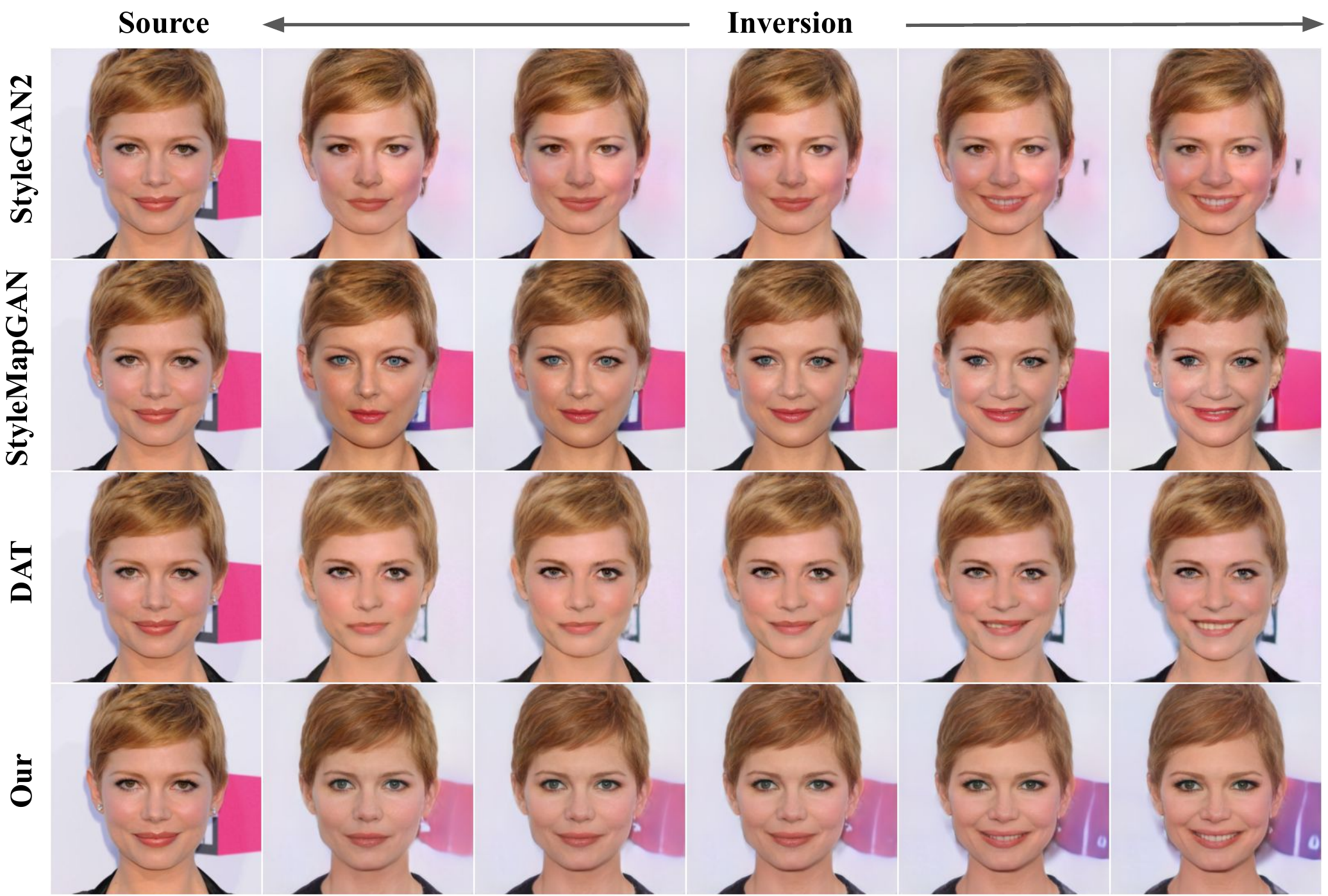}
            \caption{Smile}
    \label{fig:edit_real_smile}
        \end{subfigure}%
        \hfill
        \begin{subfigure}[b]{0.49\textwidth}
                \includegraphics[width=\linewidth]{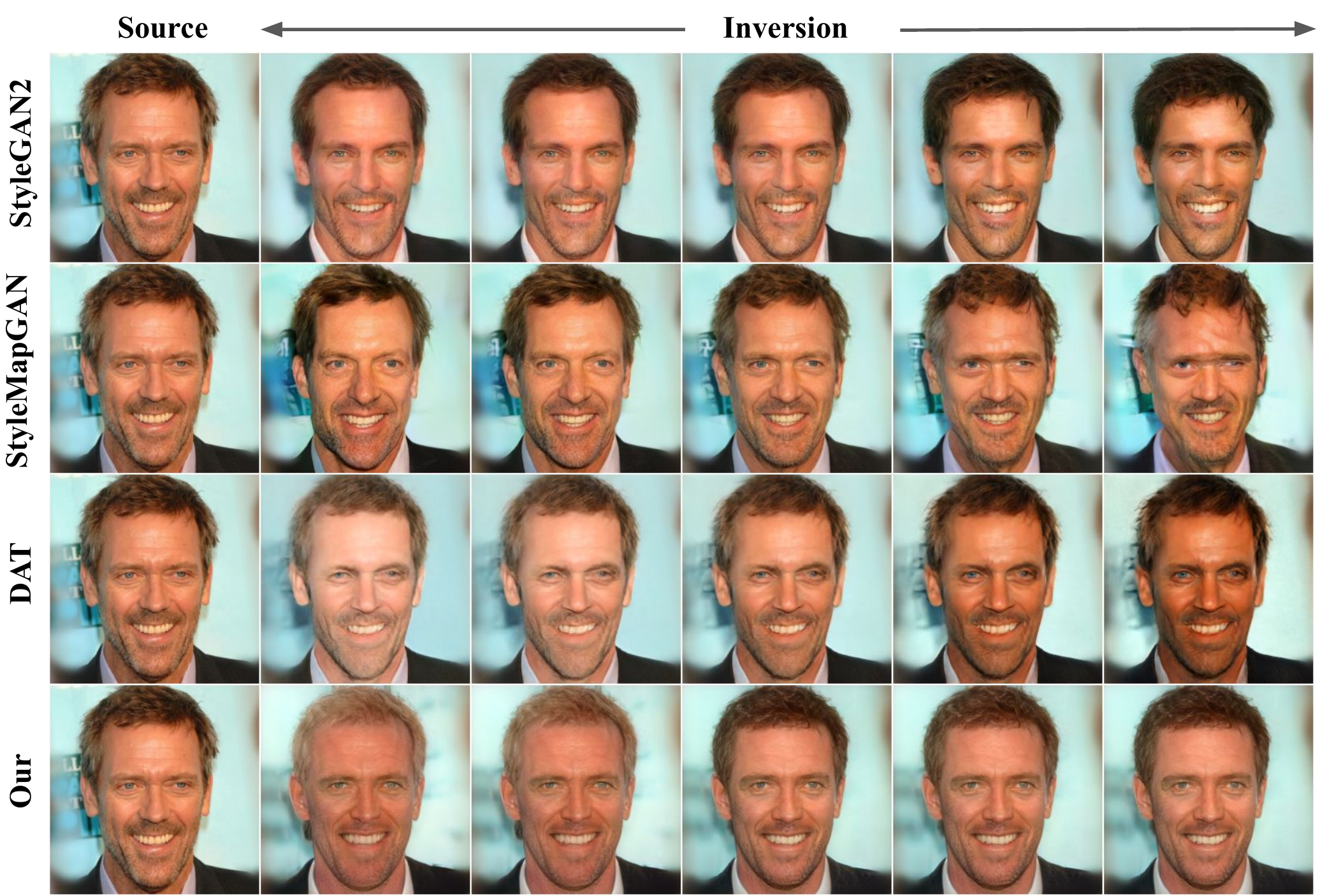}
                \caption{Black Hair}
    \label{fig:edit_real_black_hair}
        \end{subfigure}%
        \hfill
        \begin{subfigure}[b]{0.49\textwidth}
                \includegraphics[width=\linewidth]{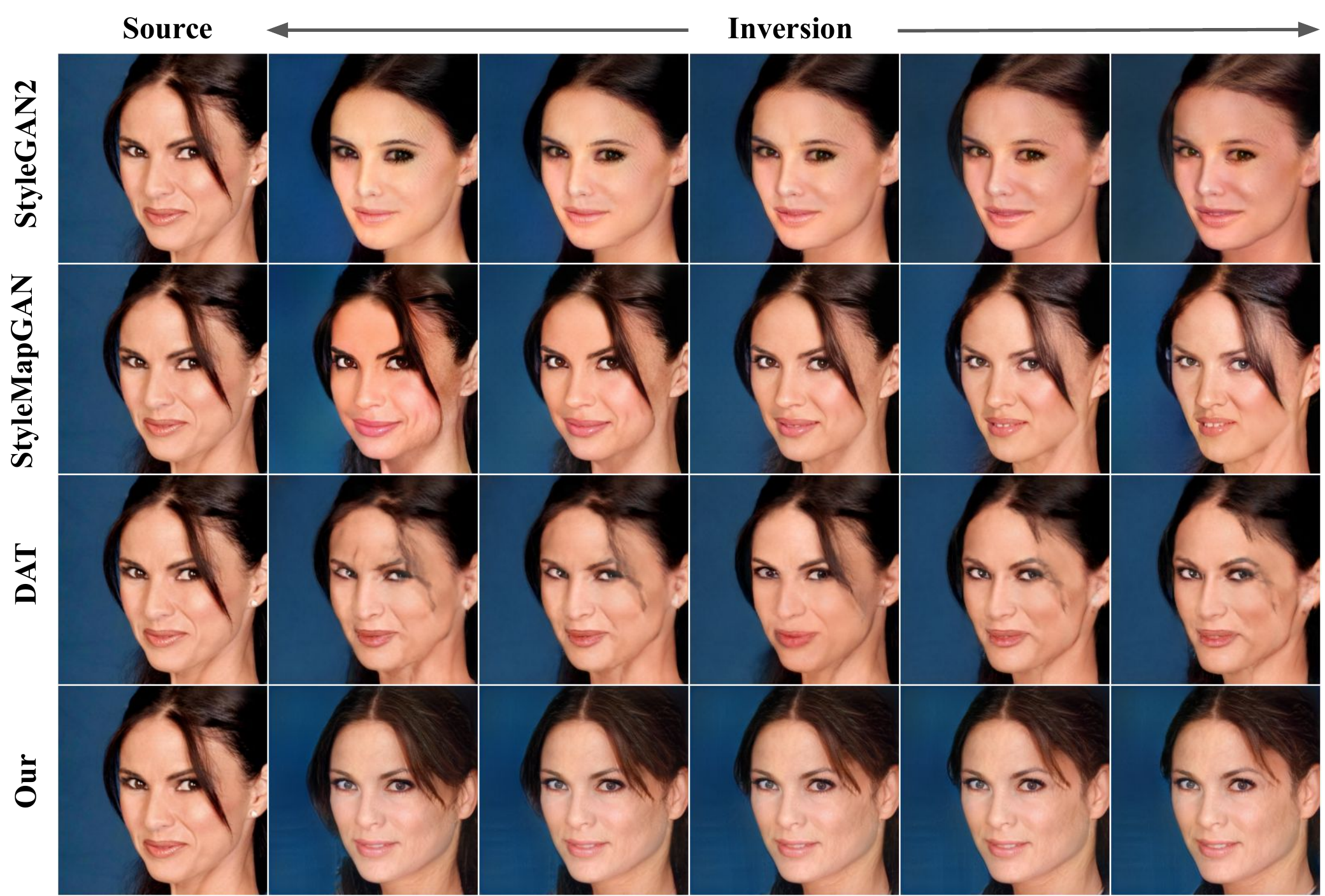}
                \caption{Bald}
    \label{fig:edit_real_bald}
        \end{subfigure}
        \vspace{-0.25cm}
        \caption{\textbf{Real Image Editing Comparison}. Images on the first column are the real source images. The fourth column shows the reconstruction results, which are semantically interpolated to the left and right sides.
        }\label{fig:real_edit_celeba}
        \vspace{-0.35cm}
\end{figure*}

\begin{figure}[]
  \centering
    \vspace{-0.18cm}
    \includegraphics[width=0.99\linewidth]{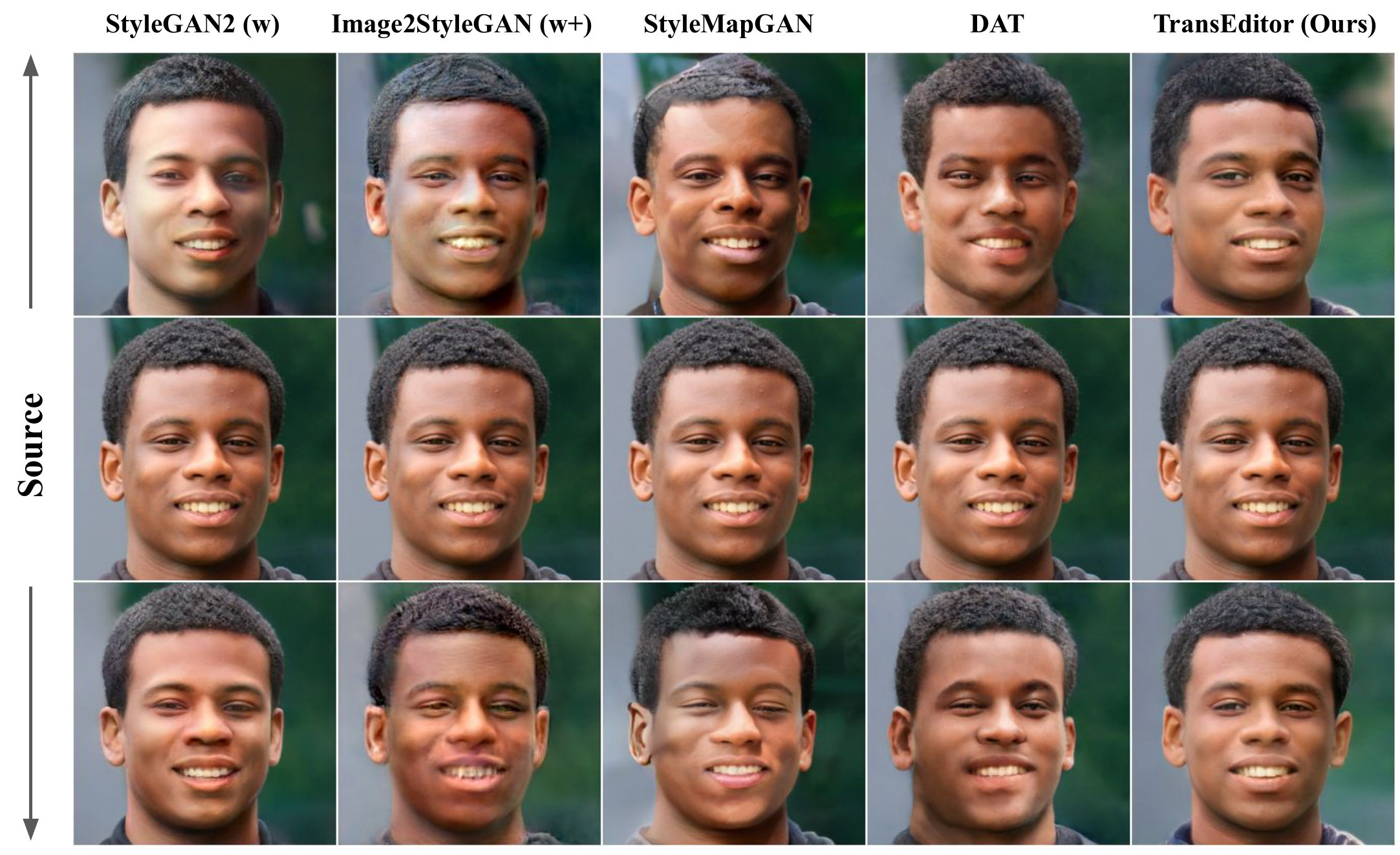}
    \vspace{-0.35cm}
    \caption{\textbf{Pose editing comparison.}}
    \vspace{-0.37cm}
   \label{fig:editing}
\end{figure}

\cref{fig:real_edit_celeba} shows more editing results of real images compared with other methods. StyleMapGAN~\cite{kim2021exploiting} is prone to face distortion when editing attributes. DAT~\cite{kwon2021diagonal} and StyleGAN2~\cite{karras2020analyzing} are prone to hue changes. Our model TransEditor achieves the best editing performance. In \cref{fig:editing}, we provide the additional comparison with w+ space (Image2StyleGAN~\cite{abdal2019image2stylegan}) using the same optimization approach for inversion. Our method still achieves the best results.

\cref{tab: Re-scoring Comparision ID} shows the results of Identity Re-scoring on complicated attributes pose and gender, compared with other methods. For DAT~\cite{kwon2021diagonal} and TransEditor, the pose is edited using content space and $\mathcal{P}$-space, respectively, since the structure information is contained in those spaces. Gender is edited using two spaces simultaneously. The result in the first row shows the identity preservation during pose editing. Our method surpasses others by a large margin. This observation is consistent with our quantitative observation on the pose editing results in the main text.  \cref{tab:FID} shows the FID metric compared with other methods.

\begin{table}
  \centering
  \small
  \caption{\textbf{Identity Re-scoring Calculation.} Compared between StyleGAN2~\cite{karras2020analyzing}, StyleMapGAN~\cite{kim2021exploiting}, DAT~\cite{kwon2021diagonal}, and the proposed TransEditor~(Ours).}
  \vspace{-0.2cm}
  \scalebox{0.95}{
  \begin{tabular}{c|c c c c }
    \toprule
     & &\textbf{ID}$\downarrow$& \\
    \midrule
     
     Method& StyleGAN2& StyleMapGAN & DAT& Ours \\
    \midrule 
    \textbf{Pose} &7.528&25.668&28.693&\textbf{2.326}\\
    \textbf{Gender} &1.240&1.209&1.323&\textbf{1.135}\\
    
    \bottomrule
  \end{tabular}}
  \label{tab: Re-scoring Comparision ID}
 \vspace{-0.25cm}
\end{table}

\subsection{More Visualization Results}

\begin{figure*}[]
  \centering
  \begin{subfigure}{0.475\linewidth}
    \includegraphics[height=9cm, width=9cm]{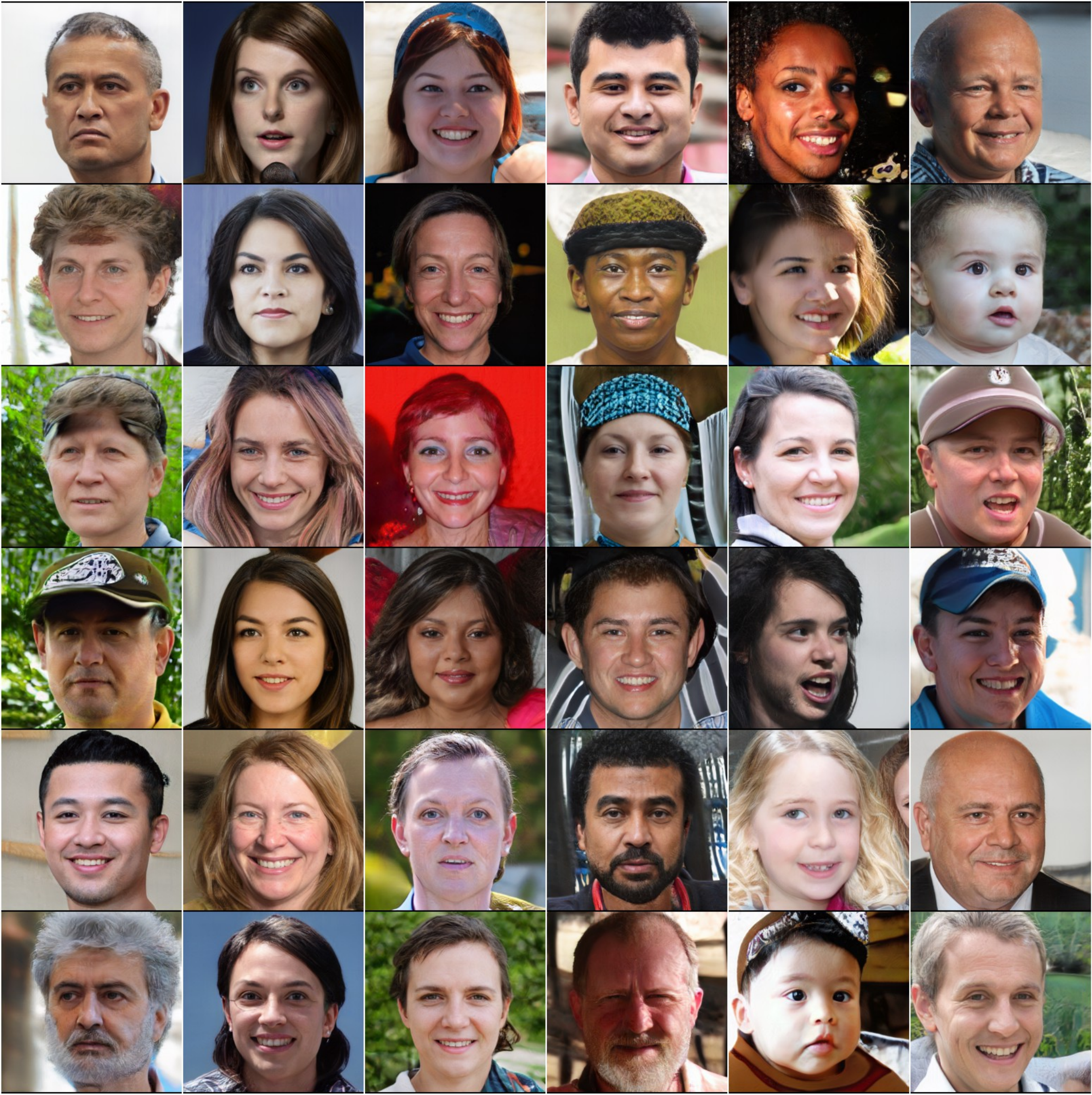} 
    \caption{Fix the $\bm{p}$ code and sample $\bm{z}$ code}
    \label{fig:short-a}
  \end{subfigure}
  \hfill
  \begin{subfigure}{0.475\linewidth}
    \includegraphics[height=9cm, width=9cm]{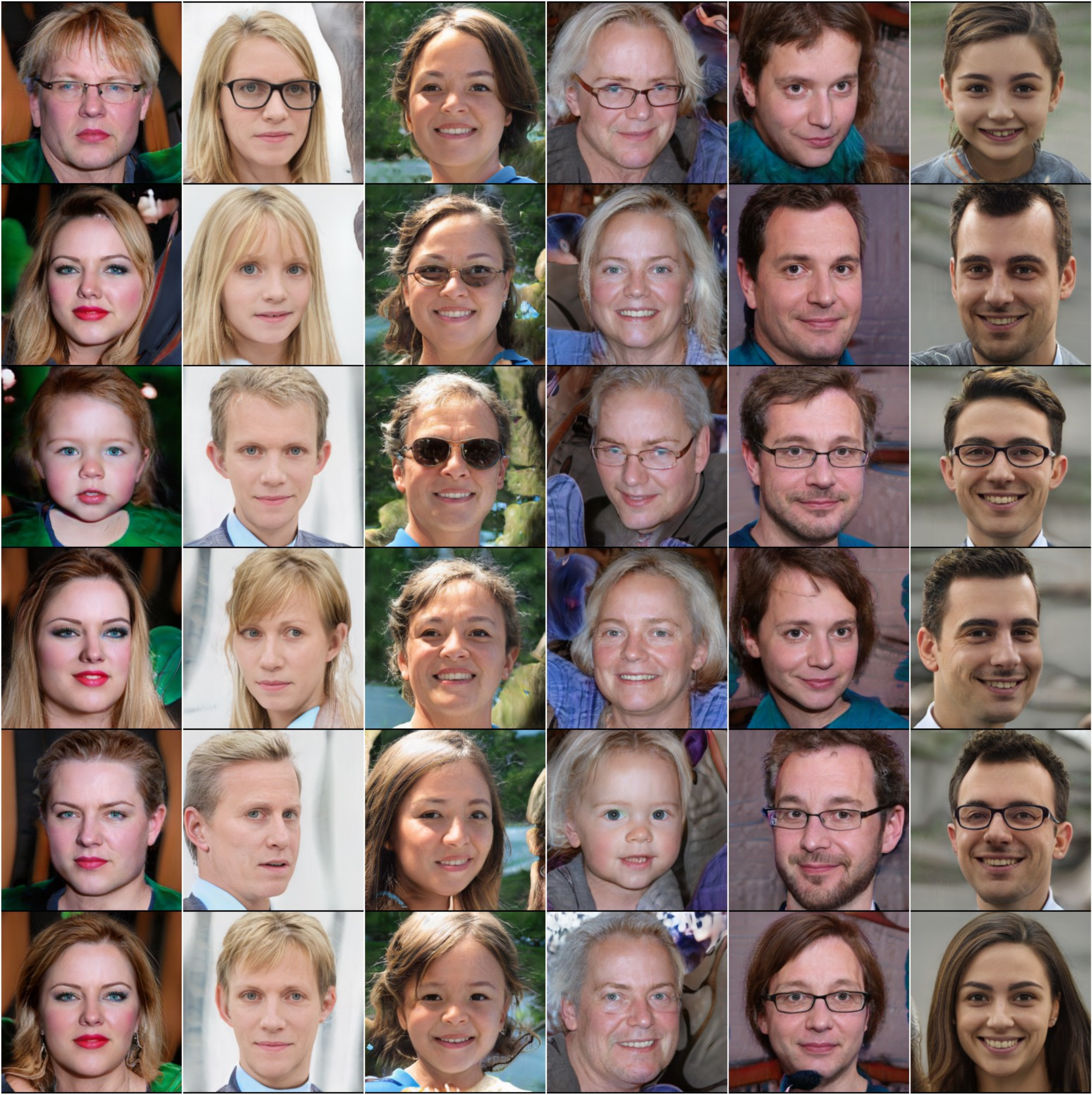}
    \caption{Fix the $\bm{z}$ code and sample $\bm{p}$ code}
    \label{fig:short-b}
  \end{subfigure}
  \caption{\textbf{Example of disentanglement of dual latent spaces on the FFHQ-256 dataset.} Each column in (a) is generated by a fixed $\bm{p}$ code and a randomly sampled $\bm{z}$ code. Note that re-sampling the $\bm{z}$ code would not influence the head pose. Similarly, for (b), each column shares the same $\bm{z}$ code. The images generated bare similar lighting, hair color, skin color. This shows the semantic disentanglement of dual latent spaces of TransEditor.}
  \label{fig:ffhq-swap}
\end{figure*}

\begin{figure*}[!h]
  \centering
  \begin{subfigure}{0.475\linewidth}
    \includegraphics[width=\linewidth
    ]{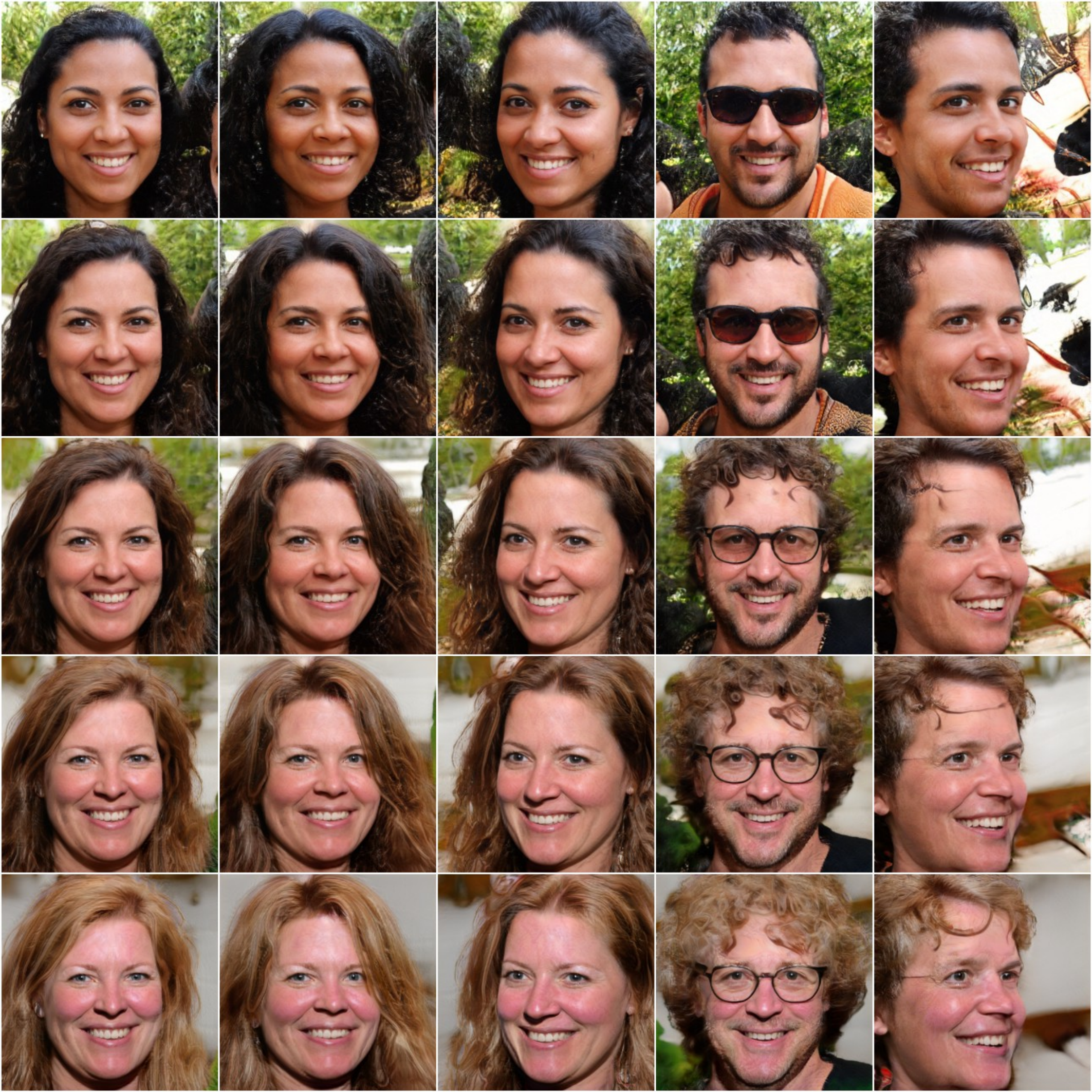} 
    \caption{Interpolate $\bm{z}$ code}
    \label{fig:short-a}
  \end{subfigure}
  \hfill
  \begin{subfigure}{0.475\linewidth}
    \includegraphics[width=\linewidth 
    ]{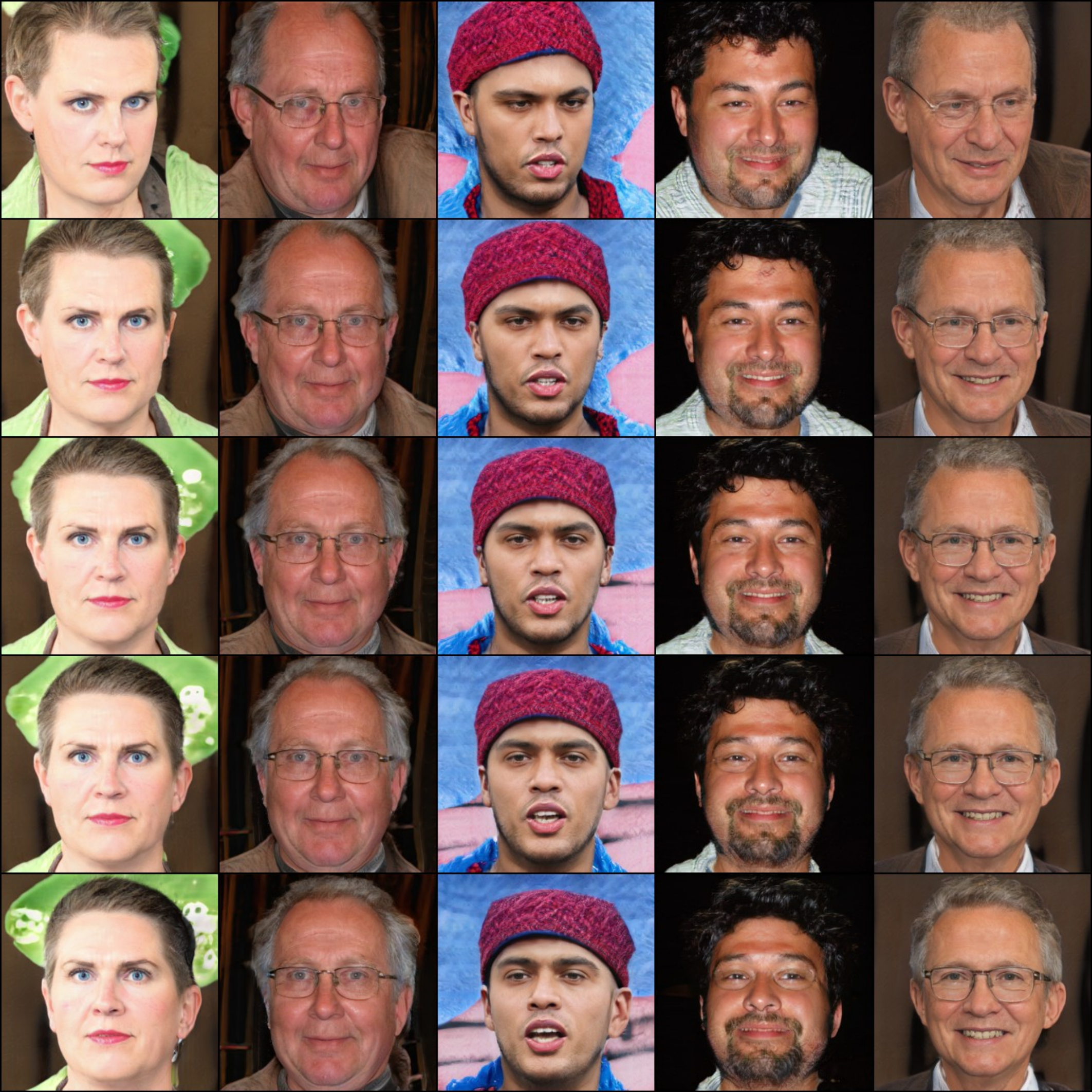}
    \caption{Interpolate $\bm{p}$ code}
    \label{fig:short-b}
  \end{subfigure}
  \caption{\textbf{Examples from interpolated dual space latent codes on the FFHQ-256 dataset.}
  In sub-figure (a), each raw has the same $\bm{z}$ code and interpolated towards the same direction. Each column has the same sampled $\bm{p}$ code.
  Notice that the interpolation of the style code gradually changes the hair color, background, and minor facial expression changes without having any effect on the person's head pose. Similarly, for the sub-figure (b), each column shares the same $\bm{z}$ code and each row shares the same $\bm{p}$ code. The interpolation of the $\bm{p}$ code changes the pose along the same direction. Notice that during interpolation, the texture information remains similar.}
  \label{fig:ffhq-interp}
\end{figure*}

Similar to the performance on the CelebA-HQ dataset~\cite{karras2017progressive}, our dual latent spaces also achieve a certain degree of semantic separation on the FFHQ dataset~\cite{karras2019style}, with $\mathcal{P}$-space controlling structural information like pose and $\mathcal{Z}$-space controlling texture information (see \cref{fig:ffhq-swap}, \cref{fig:ffhq-interp}).

\begin{figure*}[!h]
  \centering
    \includegraphics[width=\textwidth]{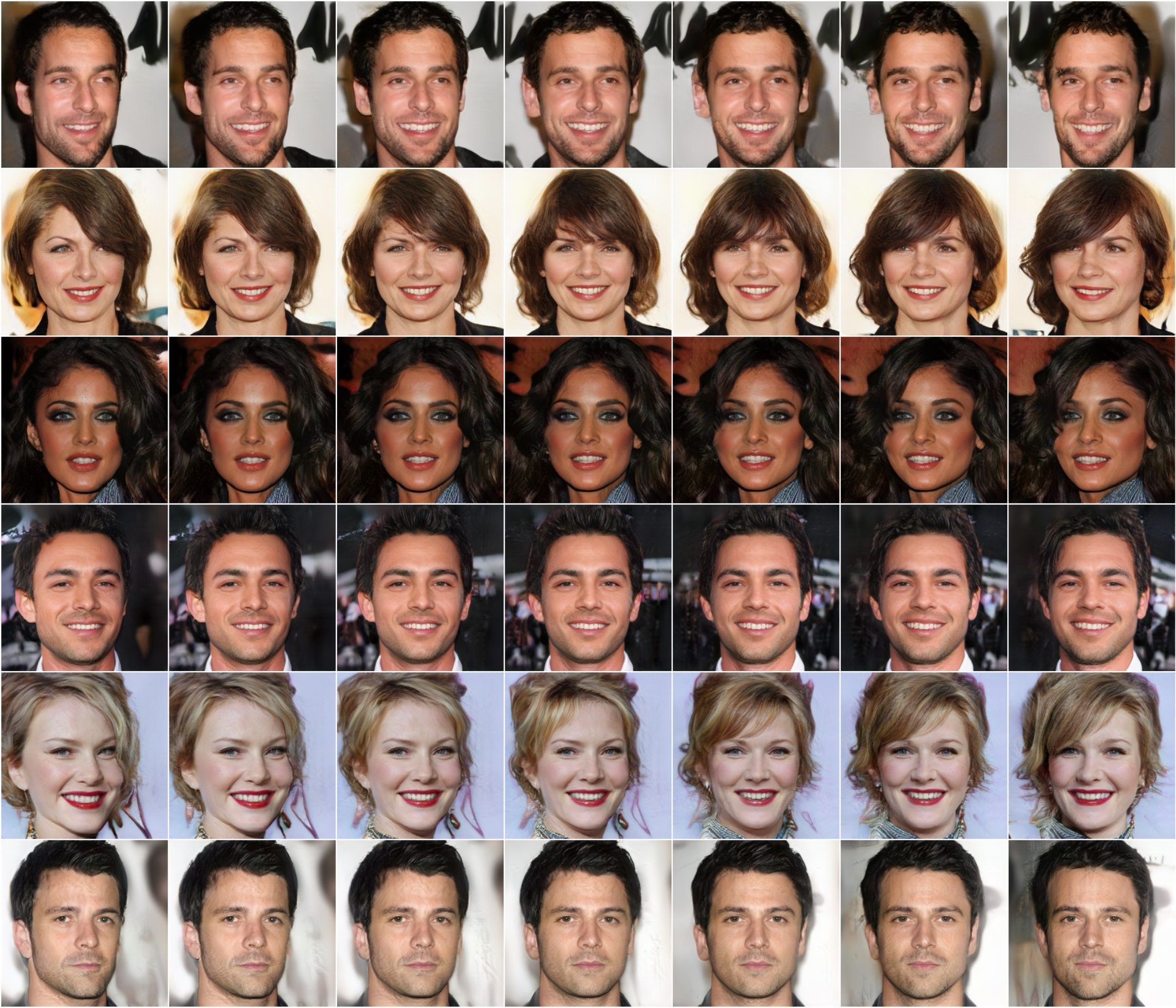}
    \caption{\textbf{Pose editing of the sampled images on CelebA-HQ-256 dataset.} Images on the fourth column are the sampled source images, which are semantically interpolated to the left and right sides. }
   \label{fig:edit_pose_celeba}
\end{figure*}

\begin{figure*}[!h]
  \centering
    \includegraphics[width=\textwidth]{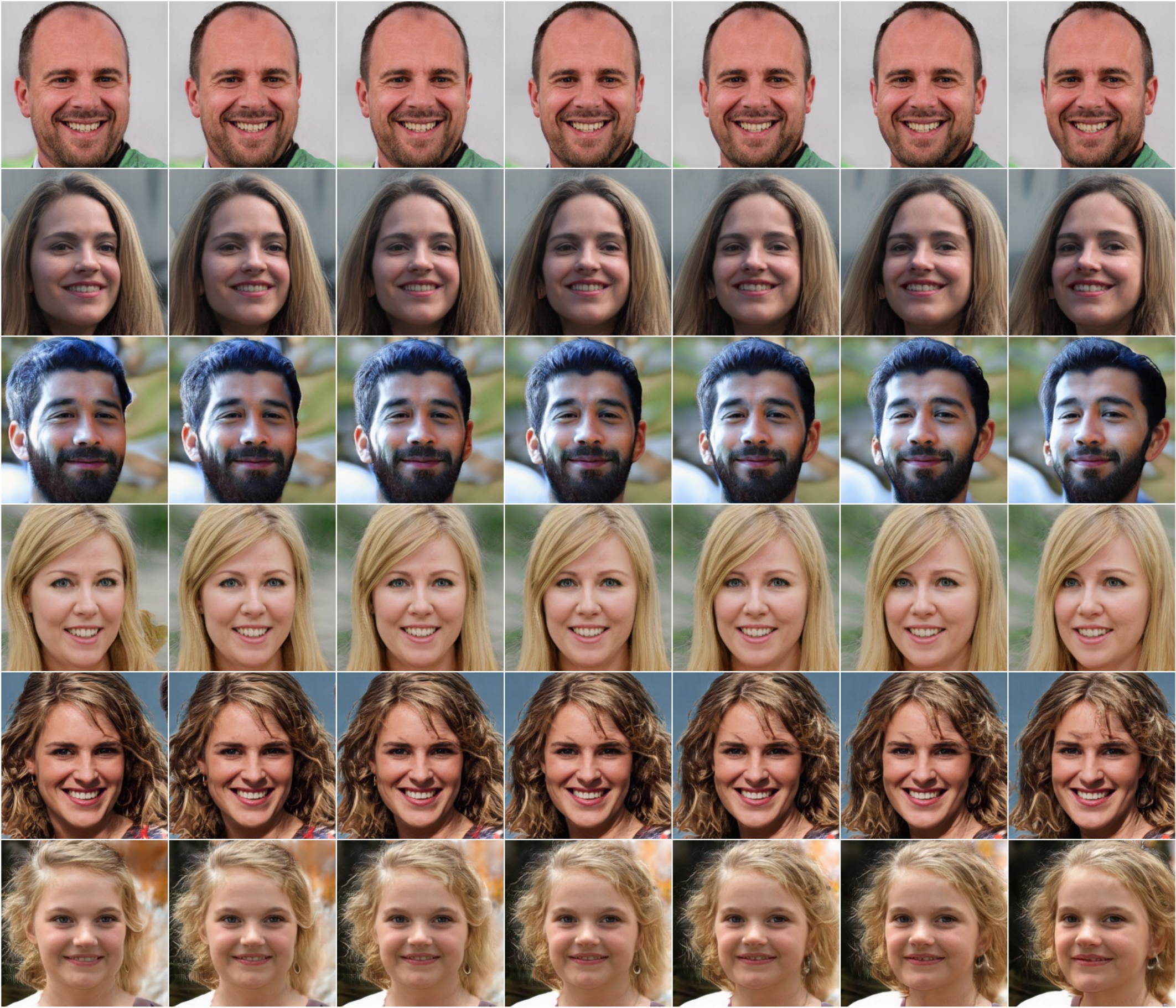}
    \caption{\textbf{Pose editing of the sampled images on FFHQ-256 dataset.} Images on the fourth column are the sampled source images, which are semantically interpolated to the left and right sides.}
   \label{fig:edit_pose_ffhq}
\end{figure*}

\begin{figure*}[!h]
  \centering
  \begin{subfigure}{0.49\linewidth}
    \includegraphics[width=\linewidth
    ]{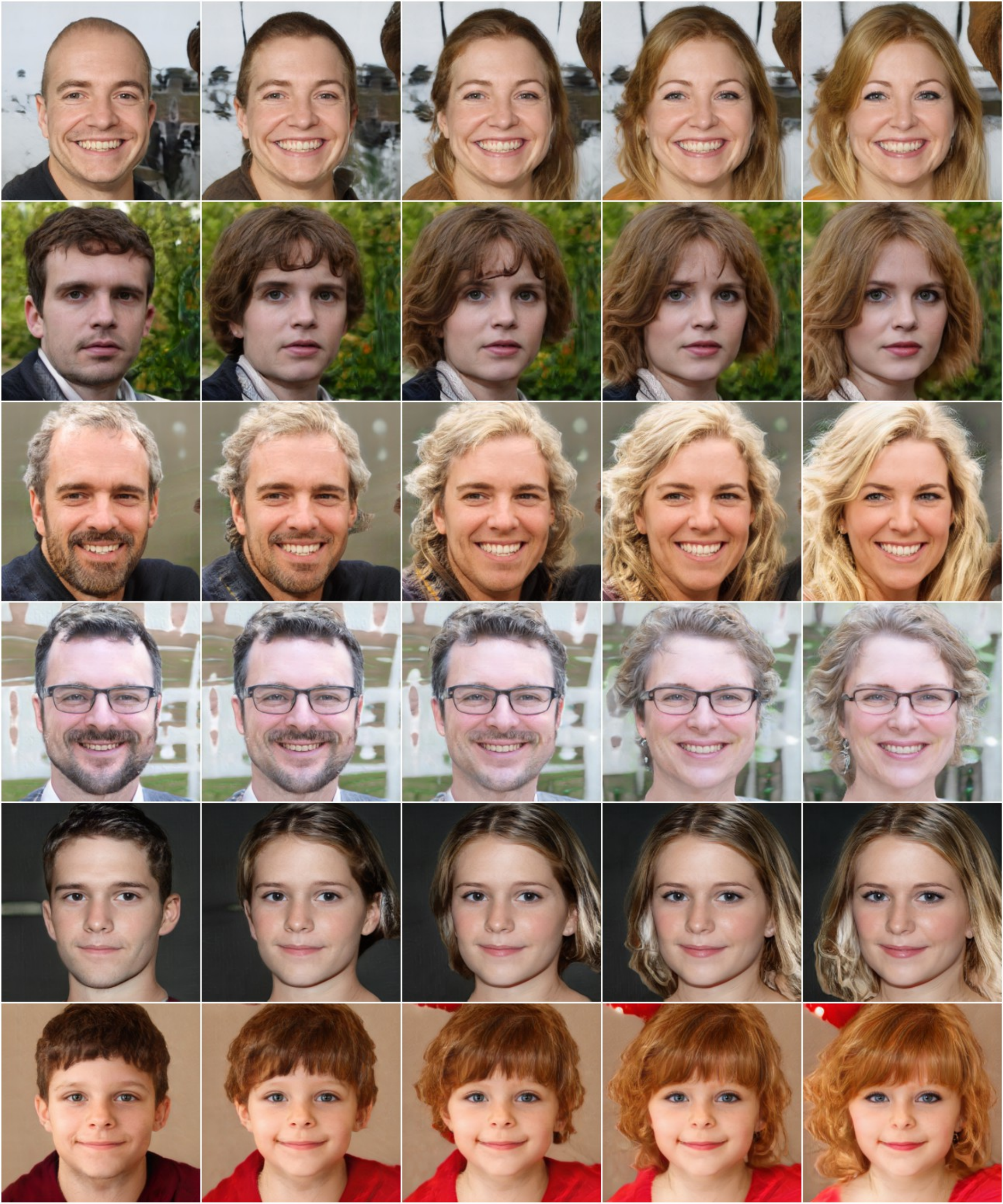} 
    \caption{Gender editing on FFHQ-256 dataset.}
    \label{fig:edit_gender_ffhq}
  \end{subfigure}
  \hfill
  \begin{subfigure}{0.49\linewidth}
    \includegraphics[width=\linewidth 
    ]{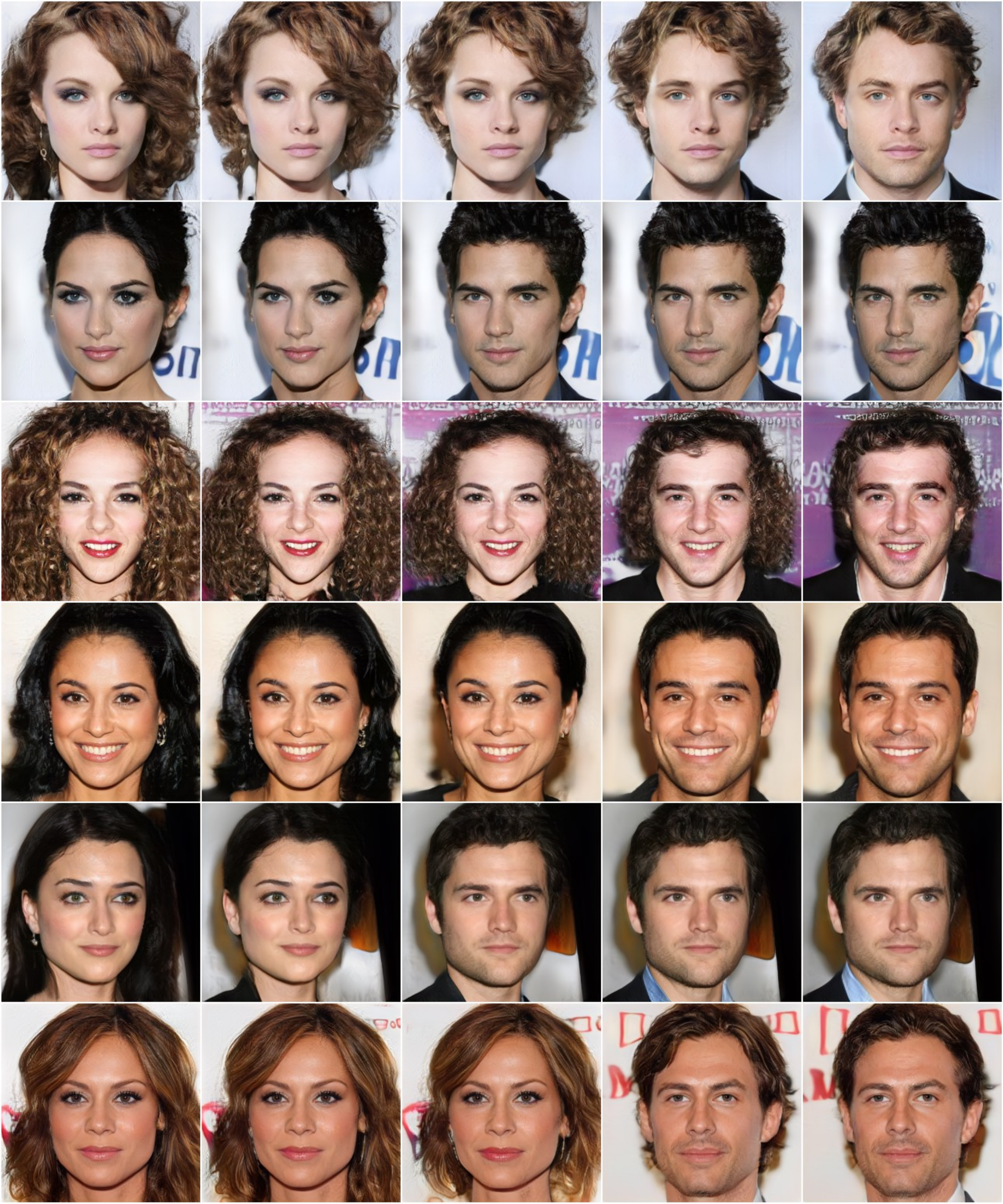}
    \caption{Gender editing on CelebA-HQ-256 dataset.}
    \label{fig:edit_gender_celeb}
  \end{subfigure}
  \caption{\textbf{Gender editing of the sampled images on FFHQ-256 dataset (a) and CelebA-HQ-256 dataset (b).} Images on the third column are the sampled source images, which are semantically interpolated to the left and right sides.}
  \label{fig:edit_gender}
\end{figure*}

\begin{figure*}[!h]
  \centering
  \begin{subfigure}{0.49\linewidth}
    \includegraphics[width=\linewidth
    ]{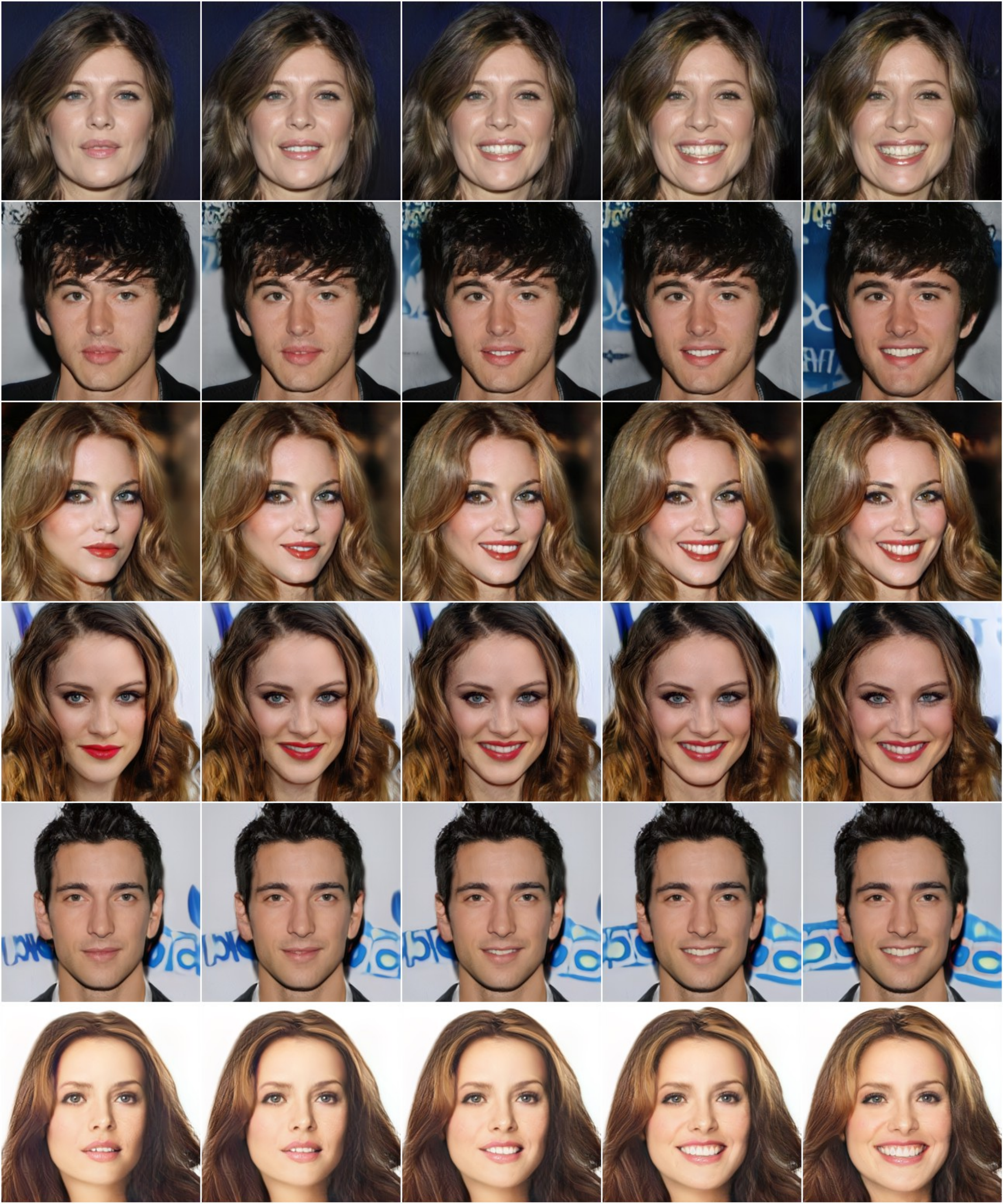} 
    \caption{Smile editing on CelebA-HQ-256 dataset.}
    \label{fig:edit_smile_celeba}
  \end{subfigure}
  \hfill
  \begin{subfigure}{0.49\linewidth}
    \includegraphics[width=\linewidth 
    ]{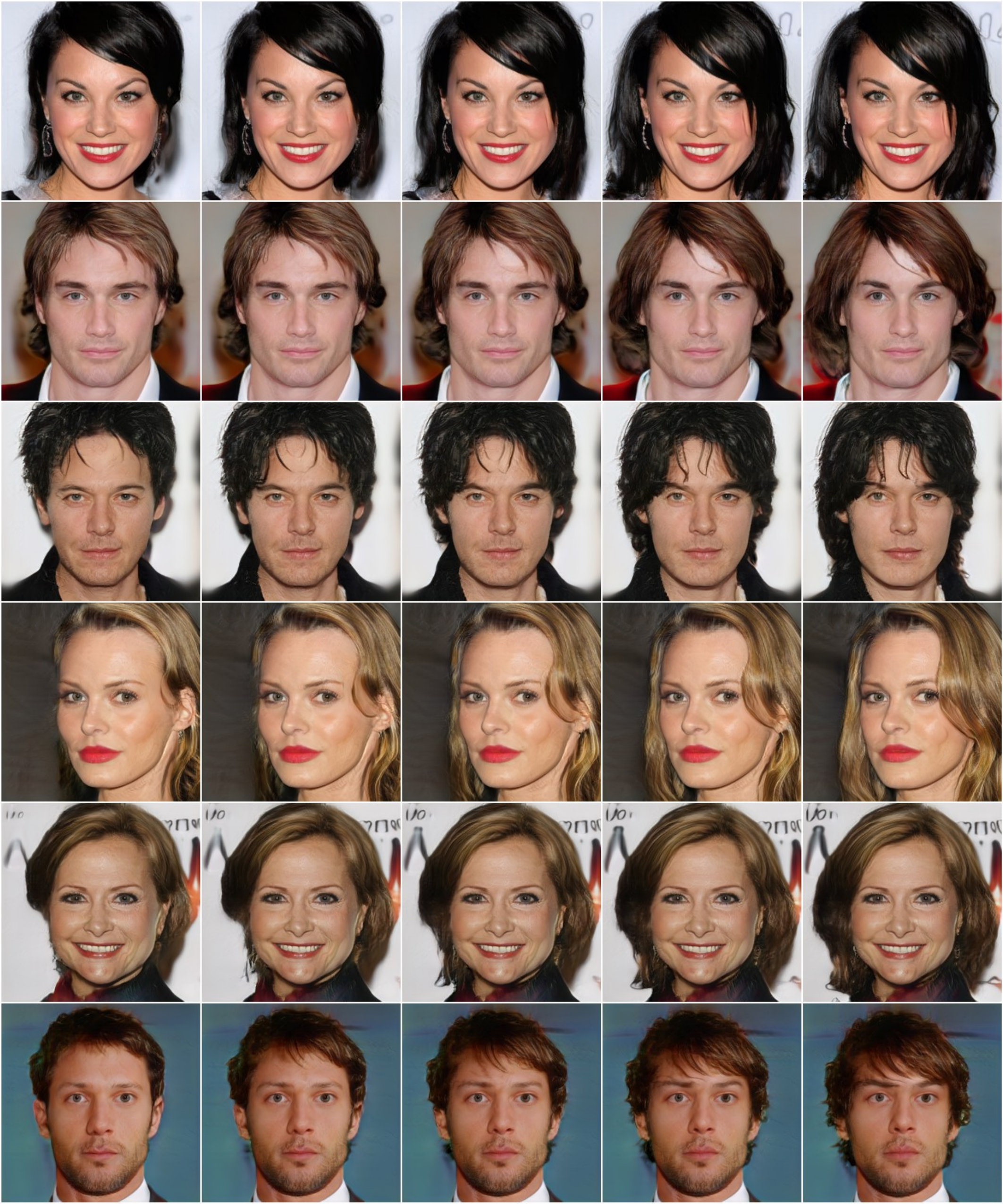}
    \caption{Wavy Hair editing on CelebA-HQ-256 dataset.}
    \label{fig:edit_wavy_hair_celeba}
  \end{subfigure}
  \caption{\textbf{Sampled Image Editing on CelebA-HQ-256 dataset.} Images on the third column are the sampled source images, which are semantically interpolated to the left and right sides.}
  \label{fig:edit_smile_wavy_celeb}
\end{figure*}

\begin{figure*}[!h]
  \centering
  \begin{subfigure}{0.49\linewidth}
    \includegraphics[width=\linewidth
    ]{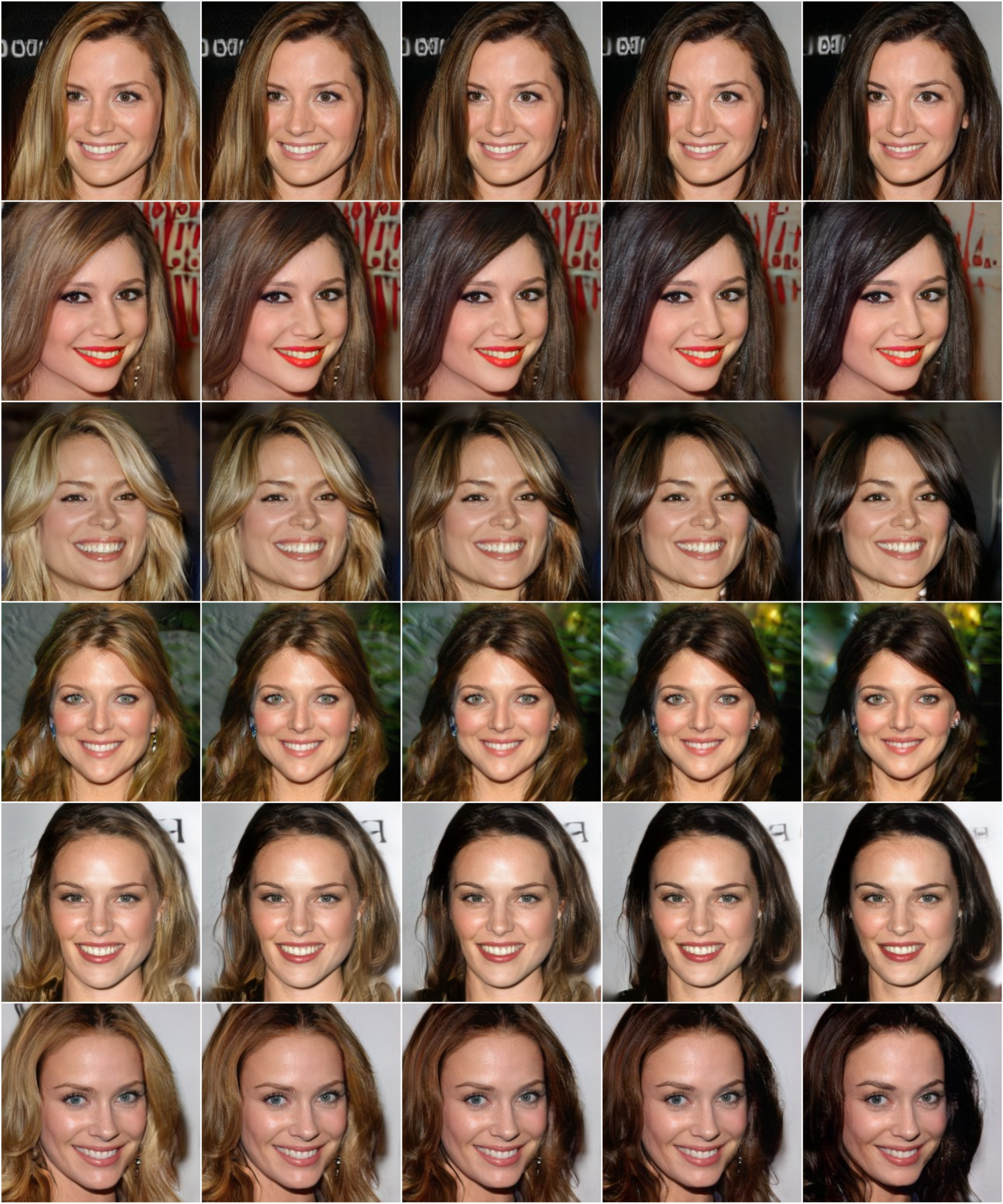} 
    \caption{Black Hair editing on CelebA-HQ-256 dataset.}
    \label{fig:edit_black_hair_celeba}
  \end{subfigure}
  \hfill
  \begin{subfigure}{0.49\linewidth}
    \includegraphics[width=\linewidth 
    ]{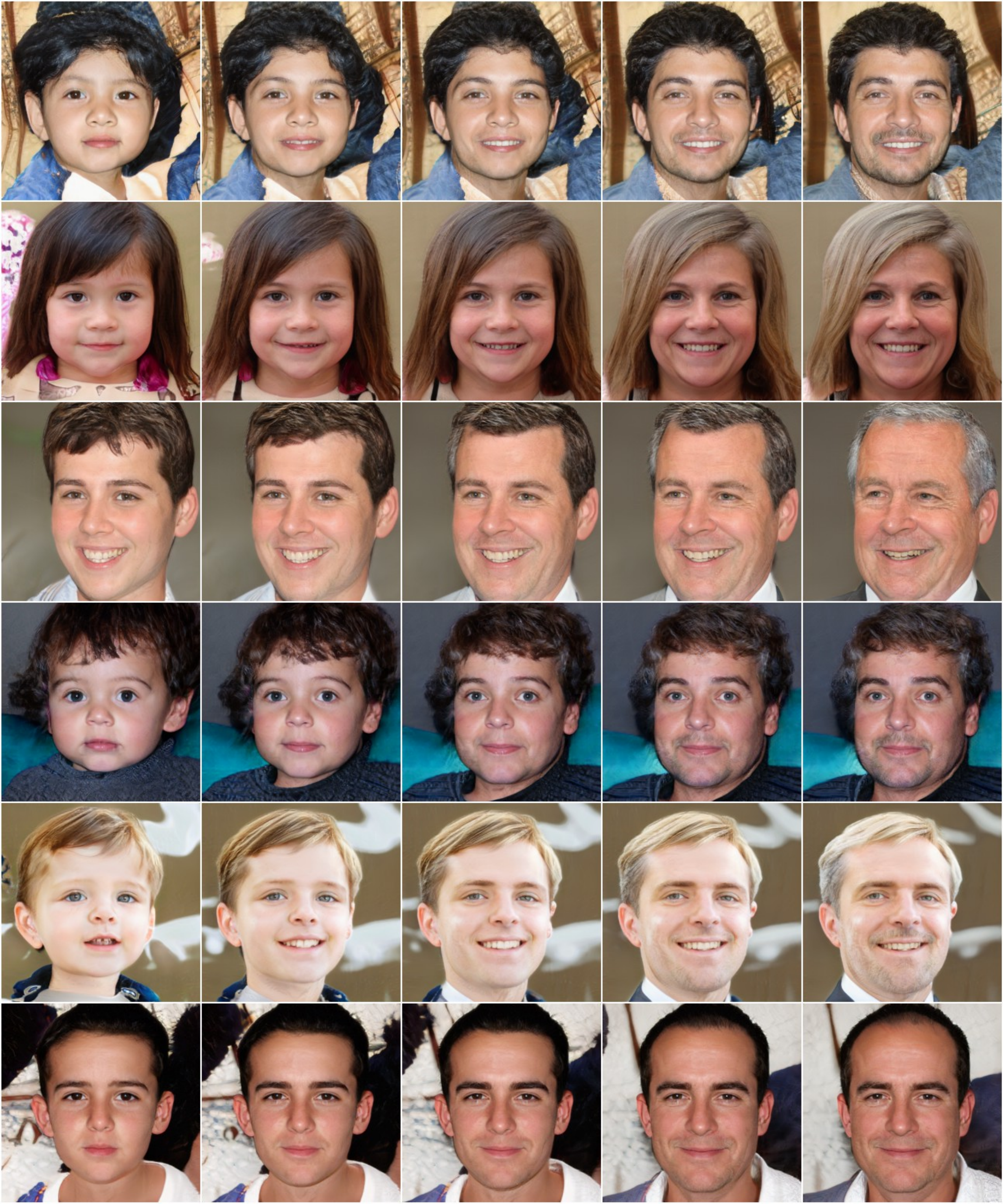}
    \caption{Age editing on FFHQ-256 dataset.}
    \label{fig:edit_age_ffhq}
  \end{subfigure}
  \caption{\textbf{Sampled image editing on CelebA-HQ-256 dataset (a) and FFHQ-256 dataset (b).} Images on the third column are the sampled source images, which are semantically interpolated to the left and right sides.}
  \label{fig:edit_age_black_hair}
\end{figure*}

The remaining figures show more editing results of TransEditor on different attributes and different datasets. \cref{fig:edit_pose_celeba} and \cref{fig:edit_pose_ffhq} are the pose editing results on the two datasets. Only $\mathcal{P}$-space is used for pose editing. Gender editing results are shown in \cref{fig:edit_gender_ffhq} and \cref{fig:edit_gender_celeb}. As mentioned in the main text, the editing of gender utilizes both spaces. \cref{fig:edit_smile_wavy_celeb} shows the smile and wavy hair editing on CelebA-HQ~\cite{karras2017progressive}, they are performed on $\mathcal{Z}$-space and $\mathcal{P}$-space respectively. \cref{fig:edit_black_hair_celeba} shows the results of black hair editing using $\mathcal{Z}$-space on CelebA-HQ~\cite{karras2017progressive}, and \cref{fig:edit_age_ffhq} shows the results of age editing on FFHQ~\cite{karras2019style}. Since change of age might involve both structure and texture variation, the editing of age is accomplished using both $\mathcal{P}$-space and $\mathcal{Z}$-space simultaneously.

\begin{table}
  \centering
  \caption{\textbf{FID Comparison.} All method are trained on FFHQ at the resolution of 256.}
  \vspace{-0.2cm}
  \scalebox{0.98}{
  \begin{tabular}{c|c}
    \toprule
    Method & $\textbf{FID}\downarrow$\\
    \midrule
    StyleGAN2~\cite{karras2020analyzing} &  4.44\\
    StyleMapGAN~\cite{kim2021exploiting} &  15.9\\
    DAT~\cite{kwon2021diagonal} & 22.50\\
    \bottomrule
    Ours & 9.32 \\
    \bottomrule
  \end{tabular}}
  \label{tab:FID}
\end{table}

\end{document}